\RequirePackage[l2tabu, orthodox]{nag}
\documentclass[12pt]{article}

\usepackage[nonatbib,final]{nips_2016}

\usepackage{setspace}
\usepackage{makecell}
\newcolumntype{M}[1]{>{\centering\arraybackslash}m{#1}}
\usepackage[T1]{fontenc}

\usepackage{tgtermes}
\usepackage{amsmath}
\usepackage{scalefnt,letltxmacro}
\LetLtxMacro{\oldtextsc}{\textsc}
\renewcommand{\textsc}[1]{\oldtextsc{\scalefont{1.10}#1}}
\usepackage[scaled=0.92]{PTSans}
\usepackage{inconsolata}
\usepackage{mathbbol}
\usepackage{amsthm}
\usepackage{amssymb}

\usepackage{setspace}

\usepackage[usenames,dvipsnames]{xcolor}
\definecolor{shadecolor}{gray}{0.9}

\usepackage[english]{babel}
\usepackage{afterpage}
\usepackage{framed}
\usepackage{nicefrac}

{\endMakeFramed}

\DeclareRobustCommand{\parhead}[1]{\textbf{#1}~}

\usepackage{lineno}

\usepackage{ragged2e}

\usepackage{marginnote}

\newcounter{parcount}

\usepackage{graphicx}
\usepackage[labelfont=bf]{caption}
\usepackage[format=hang]{subcaption}
\usepackage{wrapfig}

\usepackage{booktabs}
\usepackage{multirow}
\usepackage{longtable}

\usepackage{natbib}
\usepackage{bibunits}

\usepackage[algoruled]{algorithm2e}
\usepackage{listings}
\usepackage{fancyvrb}
\fvset{fontsize=\normalsize}

\SetKwComment{Comment}{$\triangleright$\ }{}

\usepackage[colorlinks,linktoc=all]{hyperref}
\usepackage[all]{hypcap}
\hypersetup{citecolor=Violet}
\hypersetup{linkcolor=black}
\hypersetup{urlcolor=MidnightBlue}

\usepackage[nameinlink]{cleveref}

\usepackage[acronym,smallcaps,nowarn]{glossaries}
\usepackage{listings}
\usepackage{lstbayes}
\lstdefinestyle{mystyle}{
    commentstyle=\color{OliveGreen},
    numberstyle=\tiny\color{black!60},
    stringstyle=\color{BrickRed},
    basicstyle=\ttfamily\scriptsize,
    breakatwhitespace=false,
    breaklines=true,
    captionpos=b,
    keepspaces=true,
    numbers=none,
    numbersep=5pt,
    showspaces=false,
    showstringspaces=false,
    showtabs=false,
    tabsize=2
}
\usepackage{enumitem}

\usepackage{centernot}

\crefname{lemma}{lemma}{lemmas}
\Crefname{lemma}{Lemma}{Lemmas}
\crefname{thm}{theorem}{theorems}
\Crefname{thm}{Theorem}{Theorems}
\crefname{prop}{proposition}{propositions}
\Crefname{prop}{Proposition}{Propositions}

\newtheorem{thm}{Theorem} % reset theorem numbering for each chapter
\newtheorem{prop}[thm]{Proposition}
\newtheorem{lemma}[thm]{Lemma}
\newtheorem{assumption}{Assumption}
\newtheorem{corollary}[thm]{Corollary}

\newcommand\dif{\mathop{}\!\mathrm{d}}

\newcommand{\g}{\, | \,}

\newcommand{\rmdo}{\mathrm{do}}

\newcommand{\E}[2]{\mathbb{E}_{#1}\left[#2\right]}

\usepackage{booktabs,arydshln}
\makeatletter
\def\adl@drawiv#1#2#3{%
        \hskip.5\tabcolsep
        \xleaders#3{#2.5\@tempdimb #1{1}#2.5\@tempdimb}%
                #2\z@ plus1fil minus1fil\relax
        \hskip.5\tabcolsep}
\newcommand{\cdashlinelr}[1]{%
  \noalign{\vskip\aboverulesep
           \global\let\@dashdrawstore\adl@draw
           \global\let\adl@draw\adl@drawiv}
  \cdashline{#1}
  \noalign{\global\let\adl@draw\@dashdrawstore
           \vskip\belowrulesep}}
\makeatother

\newenvironment{proofsk}{%
  \proof}{\endproof}

\newacronym{KL}{kl}{Kullback-Leibler}
\newacronym{ELBO}{elbo}{\emph{evidence lower bound}}
\newacronym{POPELBO}{pop-elbo}{\emph{population evidence lower bound}}
\newacronym{PROELBO}{pro-elbo}{\emph{profile evidence lower bound}}

\newacronym{SVI}{svi}{stochastic variational inference}

\newacronym{ADVI}{advi}{automatic differentiation variational inference}

\newacronym{GMM}{gmm}{Gaussian mixture model}
\newacronym{LDA}{lda}{latent Dirichlet allocation}
\newacronym{PF}{pf}{Poisson factorization}
\newacronym{DEF}{def}{deep exponential family}
\newacronym{RMSE}{rmse}{root mean squred error}

\newacronym{SUTVA}{sutva}{stable unit treatment value assumption}

\newacronym{PPC}{ppc}{posterior predictive check}
\newacronym{GWAS}{gwas}{genome-wide association study}
\newacronym{PPCA}{ppca}{probabilistic principal component analysis}
\newacronym{PCA}{pca}{principal component analysis}
\newacronym{KDE}{kde}{kernel density estimate}

\newacronym{LMM}{lmm}{linear mixed model}
\newacronym{LFA}{lfa}{logistic factor analysis}

\newacronym{SNP}{snp}{single-nucleotide polymorphism}
\newacronym{GLM}{glm}{generalized linear model}
\newacronym{CSI}{csi}{conditional set inference}

\usepackage{tikz}
\usetikzlibrary{bayesnet}
\usetikzlibrary{automata,arrows}
\usetikzlibrary{calc,arrows.meta,positioning}
\usepackage{pgfplots}
\pgfplotsset{compat=newest}
\pgfplotsset{plot coordinates/math parser=false}
\usepgfplotslibrary{statistics}

\pgfdeclarelayer{edgelayer}
\pgfdeclarelayer{nodelayer}
\pgfsetlayers{edgelayer,nodelayer,main}

\definecolor{hexcolor0xbfbfbf}{rgb}{0.749,0.749,0.749}

\tikzset{>=latex}
\tikzstyle{none}   = [inner sep=0pt]
\tikzstyle{line}   = [ thick, -, shorten <=1pt, shorten >=1pt ]
\tikzstyle{arrow}  = [ thick,  ->, shorten <=1pt, shorten >=1pt ]
\tikzstyle{ardash} = [ thick dotted, ->, shorten <=1pt, shorten >=1pt ]

\tikzstyle{empty}=[circle,opacity=0.0,text opacity=1.0,minimum width=4pt,minimum height=4pt]
\tikzstyle{box}=[rectangle,fill=White,draw=Black]
\tikzstyle{filled}=[circle,fill=hexcolor0xbfbfbf,draw=Black]
\tikzstyle{hollow}=[circle,fill=White,draw=Black]
\tikzstyle{param}=[rectangle,fill=Black,draw=Black,inner sep=0pt,minimum width=4pt,minimum height=4pt]
\tikzstyle{paramhollow}=[rectangle,fill=White,draw=Black,inner sep=0pt,minimum
width=4pt,minimum height=4pt]

\usepackage{environ}
\makeatletter
\newsavebox{\measure@tikzpicture}
\NewEnviron{scaletikzpicturetowidth}[1]{%
  \def\tikz@width{#1}%
  \begin{lrbox}{\measure@tikzpicture}%
  \BODY
  \end{lrbox}%
  \pgfmathparse{#1/\wd\measure@tikzpicture}%
  \BODY
}
\makeatother

\usepackage{adjustbox}

\usepackage{times}
\usepackage{adjustbox}
\usepackage{tcolorbox}
\usepackage{enumitem}
\usepackage{amssymb}
\usepackage{cuted}

\allowdisplaybreaks

\title{Multiple Causes: A Causal Graphical View}
\author{ {\bf Yixin~Wang} \\
Department of Statistics\\
Columbia University\\
\And
{\bf David M.~Blei}  \\
Data Science Institute\\
Department of Statistics and Computer Science\\
Columbia University
}

\begin{document}

\maketitle

% \begin{strip}
% \begin{tcolorbox}[width=\textwidth]
%   \centering Preliminary draft. Please do not cite or distribute.
% \end{tcolorbox}\vskip 0.3in
% \end{strip}

\begin{bibunit}[alp]

% !TEX root = dcf_SEM.tex
\begin{abstract}
  Unobserved confounding is a major hurdle for causal inference from
  observational data. Confounders---the variables that affect both the
  causes and the outcome---induce spurious non-causal correlations
  between the two. \citet{wang2018blessings} lower this hurdle with
  ``the blessings of \emph{multiple} causes,'' where the correlation
  structure of multiple causes provides indirect evidence for
  unobserved confounding. They leverage these blessings with an
  algorithm, called the deconfounder, that uses probabilistic factor
  models to correct for the confounders.  In this paper, we take a
  causal graphical view of the deconfounder. In a graph that encodes
  shared confounding, we show how the multiplicity of causes can help
  identify intervention distributions.  We then justify the
  deconfounder, showing that it makes valid inferences of the
  intervention.  Finally, we expand the class of graphs, and its
  theory, to those that include other confounders and selection
  variables.  Our results expand the theory in
  \citet{wang2018blessings}, justify the deconfounder for causal
  graphs, and extend the settings where it can be used.
\end{abstract}

%%% Local Variables:
%%% mode: latex
%%% TeX-master: "dcf_SEM"
%%% End:

% !TEX root = dcf_SEM.tex
\section{Introduction}
\label{sec:introduction}

Unobserved confounding is the major hurdle for causal inference from
observational data. Confounders are variables that affect both the
causes and the outcome.  When measured, we can account for them with
adjustments.  But when unobserved, they open back-door paths that bias
the causal inference; adjustments are not possible.

Consider the following causal problem. How does a person's diet affect
her body fat percentage?  One confounder is lifestyle: someone with a
healthy lifestyle will eat healthy foods such as boiled broccoli; but
she will also exercise frequently, which lowers her body fat.  Thus
when lifestyle is unobserved, the composition of diet will be
correlated with body fat, regardless of its true causal
effect. Compounding the difficulty, accurate measurements of lifestyle
(the confounder) are difficult to obtain, e.g., requiring expensive
real-time tracking of activities.  Lifestyle is necessarily an
unobserved confounder.~\looseness=-1

To lower the hurdle of unobserved confounding,
\citet{wang2018blessings} propose to dwell on \emph{multiple} causes.
They focus on settings where multiple causes affect a single outcome.
They found that the dependency structure of the causes can reveal
unobserved \textit{multi-cause} confounders, those that affect
multiple causes and the outcome.  They estimate those confounders with
probabilistic factor models and then use them downstream in a causal
inference.  The dependency structure of the causes is ``the blessing
of multiple causes.''

The example fits into this setting. Each type of food---broccoli,
burgers, granola bars, pizza, and so on---is a potential cause of body
fat. Further, each person's lifestyle affects multiple causes, i.e.,
their consumption of multiple types of food. People with a healthy
lifestyle eat broccoli and granola; people with an unhealthy lifestyle
eat pizza and burgers.  Thus, through the patterns of correlation
among the foods, we can infer something about each person's lifestyle,
something about the unobserved confounder.~\looseness=-1

Leveraging this idea, \citet{wang2018blessings} develop the
``deconfounder'' algorithm for causal inference. The deconfounder
constructs a random variable---called the ``substitute
confounder''---that renders the causes conditionally independent; it
then uses the substitute confounder to adjust for confounding bias. In
the Rubin causal model, they prove the deconfounder leads to unbiased
estimates of potential outcomes under the \emph{single ignorability}
assumption: each cause is conditionally independent of the potential
outcome given the observed confounders. This assumption is weaker than
the classical ignorability assumption prevalent in the potential
outcomes literature.

Here we take a causal graphical view of the blessings of multiple
causes.  What causal quantities can be identified?  How does single
ignorability translate to assumptions on causal graphs?  How does the
multiplicity of the causes resolve causal identification?  Does the
deconfounder algorithm lead to valid causal estimates on causal
graphs?  These are the questions we study.

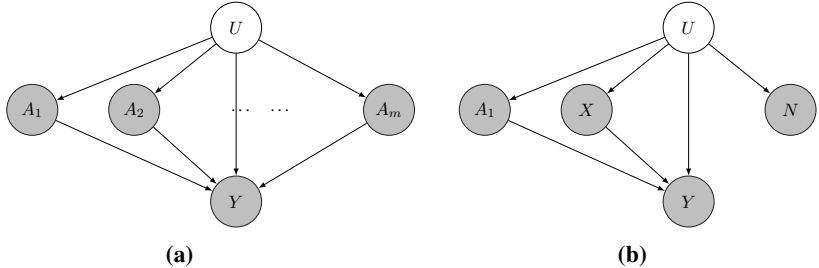
\begin{figure*}
\centering
% \begin{subfigure}[b]{0.33\textwidth}
% \centering
% \begin{adjustbox}{height=3cm}
% \input{img/twotreatfig1}
% \end{adjustbox}
% \caption{\label{fig:fig1}}
% \end{subfigure}%
\begin{subfigure}[b]{0.33\textwidth}
\centering
\begin{adjustbox}{height=3cm}
% !TEX root = ../dcf_SEM.tex
\begin{tikzpicture}
% \tikzstyle{plate} = [draw, rectangle, rounded corners, fit=#1, dashed]
[obs/.style={draw,circle,filled,minimum size=1cm}, latent/.style={draw,circle,minimum size=1cm}]
  % Define nodes
  \node[latent] (u) {$U$};
  \node[obs, below=0.6cm of u, xshift=3cm] (am) {$A_m$};
  % \node[obs, below=0.6cm of u, xshift=2cm] (am1) {$A_{m-1}$};
  \node[obs, below=0.6cm of u, xshift=-4cm] (a1) {$A_1$};
  \node[obs, below=0.6cm of u, xshift=-2cm] (a2) {$A_2$};
  \node[obs, below=2.4cm of u] (y) {$Y$};
  % Connect the nodes
  \edge{a1} {y};
  \edge{u} {a1, a2, am, y};
  \edge{a2} {y};
  \edge{am} {y};
  \node at ($(a2)!.5!(am)$) {\ldots \quad \ldots};
\end{tikzpicture}
\end{adjustbox}
\caption{\label{fig:sharedZ}}
\end{subfigure}%
\qquad\qquad
\begin{subfigure}[b]{0.33\textwidth}
\centering
\begin{adjustbox}{height=3cm}
% !TEX root = ../dcf_SEM.tex
\begin{tikzpicture}
% \tikzstyle{plate} = [draw, rectangle, rounded corners, fit=#1, dashed]
[obs/.style={draw,circle,filled,minimum size=1cm}, latent/.style={draw,circle,minimum size=1cm}]
  % Define nodes
  \node[latent] (u) {$U$};
  \node[obs, below=0.6cm of u, xshift=2cm] (n) {$N$};
  \node[obs, below=0.6cm of u, xshift=-4cm] (a1) {$A_1$};
  \node[obs, below=0.6cm of u, xshift=-2cm] (x) {$X$};
  \node[obs, below=2.4cm of u] (y) {$Y$};
  % Connect the nodes
  \edge{a1} {y};
  \edge{u} {x, y, n, a1};
  \edge{x} {y};
\end{tikzpicture}
\end{adjustbox}
\caption{\label{fig:proxy}}
\end{subfigure}%
\caption{(a) Multiple causes with shared confounding. (b) Proxy
variables for an unobserved confounder \citep{miao2018identifying}.
(Only the shaded nodes are observed.)}
\end{figure*}

Consider multiple causal inference with \textit{shared confounding}.
This setting is in the causal graph of \Cref{fig:sharedZ}, where an
unobserved confounder $U$ (lifestyle) affects multiple causes
$\{A_1, \ldots, A_m\}$ (food choices) and an outcome $Y$ (body
fat). Further consider a subset of causes $\mathcal{C}$.  We first
prove that, under suitable conditions, the intervention distribution
$p(y \g \rmdo(a_\mathcal{C}))$ is identifiable; it can be written in
terms of the observational distribution.  We then revisit the
deconfounder.  We show that it produces correct estimates of
$p(y \g \rmdo(a_\mathcal{C}))$; this result justifies the deconfounder
on causal graphs.

We then generalize the result to the larger class of graphs in
\Cref{fig:general}.  This graph contains shared confounding, measured
single-cause confounders (that only affect one cause), selection on
the unobservables, and other structures.  We prove identifiability in
this larger class as well as the correctness of the deconfounder.

\parhead{Contributions.} The main contributions of this paper are to
expand the theory of \citet{wang2018blessings} to causal graphs,
develop a new set of identification results for multiple causal
inference, prove the correctness of the deconfounder algorithm in a
new setting, and extend the deconfounder to allow for certain types of
selection bias.  These results also illustrate how the proxy variable
strategy can be put into practice. While existing identification
formulas for proxy variables involve solving integral
equations~\citep{miao2018identifying}, these results show how to
circumvent the need for such solutions by directly modeling the data.

\parhead{Related work.} This work uses and extends causal
identification with proxy variables
\citep{kuroki2014measurement,miao2018identifying}. While these works
focus on a single cause and a single outcome, we analyze multiple
causality.  We leverage multiple causes to establish causal
identification.

In more detail, proxy variables are the observable children of
unobserved confounders. Our key observation is that, in multiple
causal inference, some causes can serve as proxies for causal
identification of others.  This observation helps identify the
intervention distributions of subsets of causes; for example, the
intervention distributions of each individual cause is identifiable.
Further, unlike previous work in proxy variables, we do not need to
find two independent proxies for the unobserved confounder; some
causes themselves can serve as proxies for identifying the effect of
the other causes.

A second body of related work is on causal structural learning with
latent variables \citep{silva2006learning, mckeigue2010sparse,
  anandkumar2014tensor, frot2017learning}. These works focus on
learning whether an arrow exists between two variables, and its
direction.  We also use latent variables, but to a different end.  We
assume the direction of arrows are known and the set of causes of are
always ancestors of the outcome. Unlike the work on structure
learning, we focus on identifying and estimating the intervention
distributions of the causes.

Finally, this paper connects to the growing literature on multiple
causal inference \citep{tran2017implicit, wang2018blessings,
  ranganath2018multiple, heckerman2018accounting,
  janzing2018detecting2}. While most of these works focus on
developing algorithms, we focus here on theoretical aspects of the
problem, expanding the ideas of \citet{wang2018blessings} to
identification and estimation in causal graphs. The identification
results in this work differ from those in \citet{wang2018blessings}.
First, that work assumes a ``consistent substitute confounder,'' one
that is a deterministic function of the multiple causes; in contrast,
we allow the substitute confounder be random given the causes.
Second, we establish identification by assuming the existence of
functions of the causes that do not affect the outcome; this is a new
type of assumption in multiple causal inference.  Finally, we extend
their methods to one that can handle selection
bias~\citep{bareinboim2012controlling}, including selection that is
driven by unobserved confounders.

We note that \citet{d2019multi} provides negative examples in multiple
causal inference where some intervention distributions are not
identifiable; they also suggest collecting additional proxy variables
in multiple causal inference to resolve non-identification. The
results below do not contradict those of \citet{d2019multi}. Rather,
we focus on the intervention distributions of \emph{subsets} of the
causes; \citet{d2019multi} focuses on the intervention distributions
of \emph{all} the causes. Further, the way we use proxy variables
differs.

%%% Local Variables:
%%% mode: latex
%%% TeX-master: "dcf_SEM"
%%% End:

% !TEX root = dcf_SEM.tex

\section{Multiple causes with shared confounding}
\label{sec:sharedZ}

Consider a causal inference problem where multiple causes of interest
affect a single outcome; it is a \emph{multiple causal inference}.
Multiple causal inference deviates from classical causal inference,
where the main interest is a single cause and a single outcome.

\Cref{fig:sharedZ} provides an example. There are $m$ causes
$A_1, \ldots, A_m$ that all affect the outcome $Y$; and there is an
unobserved confounder $U$ that affects $Y$ and the causes.  This graph
exemplifies \textit{shared unobserved confounding}, where $U$ affects
multiple causes.

In this paper, the goal of multiple causal inference is to estimate
the intervention distributions on subsets of causes, $P(Y\g
\mathrm{do}(A_{\mathcal{C}}=a_\mathcal{C}))$.  It is the distribution
of the outcome $Y$ if we intervene on $A_\mathcal{C}\subset \{A_1,
\ldots, A_m\}$, which is a (strict) subset.  (For example, we might be
interested in each cause individually; then each subset contains one
of the causes.)  We will establish causal identification in this
setting and prove the validity of causal estimation with the
deconfounder algorithm. We extend these results to more general graphs
\Cref{sec:general-intervention-id}.

\subsection{Causal identification}

An intervention distribution is \emph{identifiable} if it can be
written as a function of the observed data distribution (e.g., $P(y,
a_1,\ldots, a_m)$ in \Cref{fig:sharedZ})~\citep{Pearl:2009a}.
Identifiability ensures that an intervention distribution is
\emph{estimable} from the observed data.  In \Cref{fig:sharedZ}, which
intervention distributions can be identified? In this section we prove
that, under suitable conditions, the intervention distributions of
subsets of the causes $P(y\g \mathrm{do}(a_\mathcal{C}))$ are
identifiable.\footnote{We use the abbreviation $P(y\g
\mathrm{do}(a_\mathcal{C})) 
\stackrel{\Delta}{=}P(y\g \mathrm{do}(A_\mathcal{C}=a_\mathcal{C})).$}

The starting point for causal identification with multiple causes is
the \emph{proxy variable} strategy, which focuses on causal
identification with a single
cause~\citep{kuroki2014measurement,miao2018identifying}.  Consider the
causal graph in \Cref{fig:proxy}: it has a single cause $A_1$, an
outcome $Y$, and an unobserved confounder $U$. The goal is to estimate
the intervention distribution $P(y\g \mathrm{do}(a_1))$.  There are
some other variables in the graph too.  A \emph{proxy} $X$ is an
observable child of the unobserved confounder; a \emph{null proxy} $N$
is a proxy that does not affect the outcome. The theory around proxy
variables says that the intervention distribution
$P(y\g \mathrm{do}(a_1))$ is identifiable if (1) we observe two
proxies of the unobserved confounder $U$ and (2) one of the proxies is
a null proxy~\citep{miao2018identifying}. In particular, since $N$ and
$X$ are observed, $P(y\g \mathrm{do}(a_1))$ is identifiable.

We leverage the idea of proxy variables to identify intervention
distributions in \Cref{fig:sharedZ}, multiple causes with shared
unobserved confounding.  The main idea is to use some causes as
proxies to identify the intervention distributions of other
causes. The benefit is that, with multiple causes, we do not need to
observe external proxy variables; rather the causes themselves serve
as proxies. Nor do we need to observe a null proxy, one that does not
affect the outcome (like $N$ in \Cref{fig:proxy}); we only need to
assume that there is a function of the causes that does not affect the
outcome. (We do not need to know this function either, just that at
least one such function exists.)  In short, we can use the idea of the
proxy but without collecting external data; we can work solely with
the data about the causes and the outcome. This is the ``blessing of
multiple causes'' from the causal graphical view.

We formally state the identification result.  To repeat, assume the
causal graph in \Cref{fig:sharedZ} with $m$ causes $A_{1:m}$, an
outcome $Y$, and a shared unobserved confounder $U$.  The goal is to
identify the intervention distribution of a \emph{strict subset} of
the causes $P(y\g \mathrm{do}(a_\mathcal{C}))$.

Partition the $m$ causes into three sets: $A_\mathcal{C}$ is the set
of causes on which we intervene; $A_\mathcal{X}$ is the set of causes
we use as a proxy; $A_\mathcal{N}$ is the set of causes such that
there exists a function $f(A_\mathcal{N})$ that can serve as a null
proxy.  (We discuss this assumption below.)  The latter two sets mimic
the proxy $X$ and the null proxy $N$ in the proxy variable
strategy. The sets $A_\mathcal{C}$, $A_\mathcal{X}$ and
$A_\mathcal{N}$ must be non-empty.

\begin{assumption}
  \label{assumption:null-proxy-sharedZ}
  There exists some function $f$ and a set
  $\emptyset \ne \mathcal{N} \subset \{1, \ldots,
  m\}\backslash\mathcal{C}$ such that
  \begin{enumerate}[leftmargin=*]
  \item The outcome $Y$ does not depend on $f(A_\mathcal{N})$:
    \begin{align} 
      \label{eq:nullproxyreq}
      f(A_\mathcal{N})\perp Y\g U, A_\mathcal{C}, A_\mathcal{X},
    \end{align}
    where $\mathcal{X}=  \{1, \ldots, m\}\backslash
    (\mathcal{C}\cup \mathcal{N}) \ne \emptyset.$
  \item The conditional distribution $P(u\g a_\mathcal{C},
    f(a_\mathcal{N}))$ is complete\footnote{Definition of ``complete'':
      The conditional distribution $P(u\g a_\mathcal{C}, f(a_\mathcal{N}))$
      is complete in $f(a_\mathcal{N})$  for almost all $a_\mathcal{C}$
      means for any square-integrable function $g(\cdot)$ and almost all
      $a_\mathcal{C}$,
      \[\int g(u,
        a_\mathcal{C}) P(u \g a_\mathcal{C}, f(a_\mathcal{N}))\dif u = 0
        \text{ for almost all $f(a_\mathcal{N})$ }\] 
      if and only if $g(u,
      a_\mathcal{C})= 0$ for almost all $u$.} in $f(a_\mathcal{N})$  for
    almost all $a_\mathcal{C}$. 
  \item The conditional distribution $P(f(a_\mathcal{N})\g
    a_\mathcal{C}, a_\mathcal{X})$ is complete in $a_\mathcal{X}$ for
    almost all $a_\mathcal{C}$.
  \end{enumerate}
\end{assumption}

\Cref{assumption:null-proxy-sharedZ}.1 posits that a set of causes
$A_\mathcal{N}$ exists such that some function of them
$f(A_\mathcal{N})$ can serve as a null proxy (\Cref{eq:nullproxyreq}).
Roughly, it requires $f(A_\mathcal{N})$ does not affect the outcome.
Note, it does not require that we know $f(A_{\mathcal{N}})$, just that
it exists.

When might this assumption be satisfied?  First, suppose some of the
multiple causes do not affect the outcome.  Then
\Cref{assumption:null-proxy-sharedZ}.1 reduces to the null proxy
assumption
\citep{kuroki2014measurement,miao2018identifying,d2019multi}. This
might be plausible, e.g., in a genetic study or other setting where
there are many causes.  Again, we do not need to know \textit{which}
causes are ``null causes,'' only that they exist.  Indeed, as long as
two causes are null, the theory below implies that the intervention
distributions of each individual cause is identifiable.

But this assumption goes beyond a restatement of the null proxy
assumption.  Suppose two (or more) causes only affect the outcome as a
bundle. Then the bundle can form the set $\mathcal{N}$ and the
function is one that is ``orthogonal'' to how they are combined.  As a
(silly) example, consider two of the causes to be bread and
butter. Suppose they must be served together to induce the joyfulness
of food, but not individually.  (If either is served alone, it has no
effect on joyfulness one way or the other.)  Then the function
$f(A_{\mathcal{N}})$ is XOR of the bundle; the quantity (bread XOR
butter) does not affect $Y$.  Again, the function and set must exist;
we do not need to know them.

As a more serious example, consider that HDL cholesterol, LDL
cholesterol, and triglycerides (TG) affect the risk of a heart attack
through the ratios HDL/LDL and
TG/HDL~\citep{millan2009lipoprotein}. Then HDL$\times$LDL and
TG$\times$HDL are both examples of $f(A_\mathcal{N})$ that do not
affect $Y$.  The existence of one of them suffices for
\Cref{assumption:null-proxy-sharedZ}.1.  (We discuss this assumption
in more technical detail in \Cref{sec:example-null-proxy}.)

\Cref{assumption:null-proxy-sharedZ}.2 and
\Cref{assumption:null-proxy-sharedZ}.3 are two completeness conditions
on the true causal model; they are required by the proxy variable
strategy (e.g. Conditions 2 and 3 of \citet{miao2018identifying}).
Roughly, they require that the distributions of $U$ corresponding to
different values of $f(A_{\mathcal{N}})$ are distinct; the
distributions of $f(A_{\mathcal{N}})$ relative to different
$A_{\mathcal{X}}$ values are also distinct. The two assumptions are
satisfied when we work with a causal model that satisfies the
completeness condition.

Many common models satisfy the completeness condition.
Examples include exponential families \citep{newey2003instrumental},
location-scale families \citep{hu2018nonparametric}, and nonparametric
regression models \citep{darolles2011nonparametric}. Completeness is a
common assumption posited in nonparametric causal identification
\citep{miao2018identifying, yang2017identification,
  d2011completeness}; it is often used to guarantee the existence and
the uniqueness of solutions to integral equations. See
\citet{chen2014local} for a detailed discussion of completeness.

Under \Cref{assumption:null-proxy-sharedZ}, we can identify the
intervention distribution of the subset of the causes $A_\mathcal{C}$.

\begin{thm} (Causal identification under shared confounding)
\label{thm:idsharedZ}
Assume the causal graph \Cref{fig:sharedZ}. (Note the data does not
need to be ``faithful'' to the graph---some edges can be missing.)
Under \Cref{assumption:null-proxy-sharedZ}, the intervention
distribution of the causes $A_\mathcal{C}$ is identifiable:
\begin{align}
P(y\g \mathrm{do}(a_\mathcal{C})) = \int h(y, a_\mathcal{C}, a_\mathcal{X}) P(a_\mathcal{X}) \dif a_\mathcal{X}
\end{align}
for any solution $h$ to the integral equation
\begin{align}
\label{eq:integral}
P(y\g a_\mathcal{C}, f(a_\mathcal{N})) = \int h(y, a_\mathcal{C},
a_\mathcal{X}) P(a_\mathcal{X}\g a_\mathcal{C}, f(a_\mathcal{N}))\dif
a_\mathcal{X}.
\end{align}
Moreover, the solution to \Cref{eq:integral} always exists under weak
regularity conditions in \Cref{sec:integral-sol-exist}.
\end{thm}

\begin{proofsk}
  The proof relies on the partition of the $m$ causes: $A_\mathcal{C}$
  as the causes, $A_\mathcal{X}$ as the proxies, and $A_\mathcal{N}$
  such that $f(A_\mathcal{N})$ can be a null proxy.  We then follow
  the proxy variable strategy to identify the intervention
  distributions of $A_\mathcal{C}$ using $A_\mathcal{X}$ as a proxy
  and $f(A_\mathcal{N})$ as a null proxy. We no longer have a null
  proxy like $N$ as in \Cref{fig:proxy}; all the $m$ causes can affect
  the outcome. However, \Cref{assumption:null-proxy-sharedZ}.1 allows
  $f(A_\mathcal{N})$ to play the role of a null proxy. The full proof
  is in \Cref{subsec:proof-idsharedZ}.
\end{proofsk}

\Cref{thm:idsharedZ} identifies the intervention distributions of
subsets of the causes $A_\mathcal{C}$; it writes $P(y\g
\mathrm{do}(a_\mathcal{C}))$ as a function of the observed data
distribution $P(y, a_\mathcal{C}, a_\mathcal{X}, a_\mathcal{N})$. In
particular, it lets us identify the intervention distributions of
individual causes $P(y\g
\mathrm{do}(a_i)), i=1, \ldots, m$. By using the causes themselves as
proxies, \Cref{thm:idsharedZ} exemplifies how the multiplicity of the
causes enables causal identification under shared unobserved
confounding.

\subsection{Causal estimation with the deconfounder}
\label{subsec:dcf-sharedZ}

\Cref{thm:idsharedZ} guarantees that the intervention distribution
$P(y\g
\mathrm{do}(a_\mathcal{C}))$ is estimable from the observed data.
However, it involves solving an integral equation
(\Cref{eq:integral}). This integral equation is hard to solve except
in the simplest linear Gaussian case \citep{carrasco2007linear}. How
can we estimate $P(y\g \mathrm{do}(a_\mathcal{C}))$ in practice?

We revisit the deconfounder algorithm in \citet{wang2018blessings}. We
show that the deconfounder correctly estimates the intervention
distribution $P(y\g \mathrm{do}(a_\mathcal{C}))$; it implicitly solves
the integral equation in \Cref{eq:integral} by modeling the data. This
result justifies the deconfounder from a causal graphical perspective.

\parhead{The deconfounder algorithm.} We first review the algorithm.
Given the causes $A_1, \ldots, A_m$ and the outcome $Y$, the
deconfounder proceeds in three steps:
\begin{enumerate}[leftmargin=*]
\item \textbf{Construct a substitute confounder.} Based \emph{only} on
  the (observed) causes $A_1, \ldots, A_m$, it first constructs a
  random variable $\hat{Z}$ such that all the causes are conditionally
  independent:
\begin{align}
  \label{eq:dcfreq1}
  \hat{P}(a_1,\ldots, a_m, \hat{z}) =
  \hat{P}(\hat{z})\prod^{m}_{j=1}\hat{P}(a_j\g
  \hat{z}),
\end{align}
where $\hat{P}(\cdot)$ is consistent with the observed data $P(a_1,
\ldots, a_m) = \int \hat{P}(a_1,\ldots, a_m, \hat{z})
\dif \hat{z}.$
The random variable $\hat{Z}$ is called a \emph{substitute
  confounder}; it does not necessarily coincide with the unobserved
confounder $U$. The substitute is constructed using probabilistic
models with local and global
variables~\citep{tipping1999probabilistic} (e.g., mixture models,
matrix factorization, topic models, and others).

\item \textbf{Fit an outcome model.} The next step is to estimate how
  the outcome depends on the causes and the substitute confounder
  $\hat{P}(y\g a_1, \ldots, a_m, \hat{z})$. This \textit{outcome
    model} is fit to be consistent with the observed data:
  \begin{align}
    \label{eq:dcfreq2}
    P(y, a_1, \ldots, a_m)
    =\int \hat{P}(y\g a_1, \ldots, a_m,
    \hat{z}) \hat{P}(a_1,\ldots, a_m, \hat{z})\dif \hat{z}.
  \end{align}
  Along with the first step, the deconfounder gives the joint distribution
  $\hat{P}(y, a_1, \ldots, a_m, \hat{z})$.

\item \textbf{Estimate the intervention distribution.} The final step
  estimates the intervention distribution
  $P(y\g \mathrm{do}(a_\mathcal{C}))$ by integrating out the
  non-intervened causes and the substitute confounder,
  \begin{align}
    \label{eq:dcf}
    \hat{P}(y\g \mathrm{do}(a_\mathcal{C})) \stackrel{\Delta}{=}\int \hat{P}(y\g a_1, \ldots, a_m,
    \hat{z})
    \times \hat{P}(a_{\{1, \ldots, m\}\backslash\mathcal{C}} ,
    \hat{z})\dif \hat{z}\dif a_{\{1, \ldots,
    m\}\backslash\mathcal{C}}.
  \end{align}
  This is the estimate of the deconfounder.
\end{enumerate}

\parhead{The correctness of the deconfounder.}  Note that many
possible $\hat{P}(\cdot)$'s satisfy the deconfounder requirements
(\Cref{eq:dcfreq1,eq:dcfreq2}); the algorithm outputs one such
$\hat{P}$. Under suitable conditions, we show that any such $\hat{P}$
provides the correct causal estimate
$P(y\g \mathrm{do}(a_\mathcal{C}))$.

\begin{assumption}
\label{assumption:dcf-sharedZ}
The deconfounder estimate $\hat{P}(y, a_1, \ldots, a_m, \hat{z})$
satisfies two conditions:
\begin{enumerate}[leftmargin=*, noitemsep]
\item It is consistent with \Cref{assumption:null-proxy-sharedZ}.1,
$\hat{P}(y\g a_\mathcal{C}, a_\mathcal{X}, f(a_\mathcal{N}), \hat{z}) = \hat{P}(y\g a_\mathcal{C}, a_\mathcal{X}, \hat{z})$.
\item The conditional distribution
  $\hat{P}(\hat{z}\g a_\mathcal{C}, a_\mathcal{X})$ is complete in
  $a_\mathcal{X}$ for almost all $a_\mathcal{C}$.
\end{enumerate}
\end{assumption}

\Cref{assumption:dcf-sharedZ}.1 roughly requires that there exists a
function $f$ and a subset of the causes $A_\mathcal{N}$ such that
$f(A_\mathcal{N})$ does not affect the outcome in the deconfounder
outcome model. (When the number of causes goes to infinity,
\Cref{assumption:dcf-sharedZ}.1 reduces to
\Cref{assumption:null-proxy-sharedZ}.1.) We emphasize that
$f(A_\mathcal{N})$ is not involved in calculating the deconfounder
estimate (\Cref{eq:dcf}); it only appears in
\Cref{assumption:dcf-sharedZ}.1. Hence the correctness of the
deconfounder does not require specifying $f(\cdot)$ and
$A_\mathcal{N}$, just that it exists.

\Cref{assumption:dcf-sharedZ}.2 requires that the distributions of
$\hat{Z}$ corresponding to different values of $A_\mathcal{X}$ are
distinct. It is a similar completeness condition as in \Cref{assumption:null-proxy-sharedZ}.

Now we state the correctness result of the deconfounder.
\begin{thm} (Correctness of the deconfounder under shared confounding)
\label{thm:dcfcorrect-sharedZ}
Assume the causal graph \Cref{fig:sharedZ}. Under
\Cref{assumption:null-proxy-sharedZ}, \Cref{assumption:dcf-sharedZ}
and weak regularity conditions, the deconfounder provides correct
estimates of the intervention distribution:
\begin{align}
\hat{P}(y\g \mathrm{do}(a_\mathcal{C})) = P(y\g \mathrm{do}(a_\mathcal{C})),
\end{align}
where $\hat{P}(y\g \mathrm{do}(a_\mathcal{C}))$ is computed from
\Cref{eq:dcf}.
\end{thm}

\begin{proofsk}
The proof of \Cref{thm:dcfcorrect-sharedZ} relies on a key
observation: the deconfounder implicitly solves the integral equation
(\Cref{eq:integral}) by modeling the observed data with $\hat{P}(y,
a_1, \ldots, a_m, \hat{z})$. \Cref{assumption:dcf-sharedZ}.2
guarantees that the deconfounder estimate can be written as
\begin{align}
\label{eq:dcf2}
\hat{P}(y\g a_\mathcal{C},
\hat{z}) = \int \hat{h}(y, a_\mathcal{C},
a_\mathcal{X}) \hat{P}(a_\mathcal{X}\g \hat{z})\dif a_\mathcal{X}
\end{align}
under weak regularity conditions; this function $\hat{h}(y,
a_\mathcal{C}, a_\mathcal{X})$ also solves the integral equation
(\Cref{eq:integral}). The deconfounder uses this solution to form an
estimate of $P(y\g \mathrm{do}(a_\mathcal{C}))$; this estimate is
correct because of \Cref{thm:idsharedZ}. The full proof is in
\Cref{subsec:proof-dcfcorrect}.
\end{proofsk}

\Cref{thm:dcfcorrect-sharedZ} justifies the deconfounder for multiple
causal inference under shared confounding (\Cref{fig:sharedZ}). It
proves that the deconfounder correctly estimates the intervention
distributions when they are identifiable.  This result complements
Theorems 6--8 of \citet{wang2018blessings}; it establishes
identification and correctness by assuming there exists some function
of the causes that does not affect the outcome. In contrast, Theorems
6--8 of \citet{wang2018blessings} assume a ``consistent substitute
confounder,'' that the substitute confounder is a deterministic
function of the causes. Their assumption is stronger; conditional on
the causes, \Cref{thm:dcfcorrect-sharedZ,thm:idsharedZ} allow the
substitute confounder to be random.

\Cref{thm:dcfcorrect-sharedZ} also shows that we can leverage the
deconfounder algorithm to put the proxy variable strategy into
practice. While existing identification formulas of proxy variables
involves solving integral equations \citep{miao2018identifying},
\Cref{thm:dcfcorrect-sharedZ} shows how to circumvent this need by
directly modeling the data and applying the deconfounder; it
implicitly solves the integral equations with the modeled data.

To illustrate \Cref{thm:idsharedZ,thm:dcfcorrect-sharedZ},
\Cref{subsec:examplecontd} gives a linear example.

%%% Local Variables:
%%% mode: latex
%%% TeX-master: "dcf_SEM"
%%% End:

% !TEX root = dcf_SEM.tex

\section{Multiple causes on general causal graphs}
\label{sec:general-intervention-id}

We discussed causal identification and estimation when multiple causes
share the same unobserved confounder. We now extend these results to
more general causal graphs, those with several types of nodes and, in
particular, that include a selection
variable~\citep{bareinboim2014recovering}.
Using the results in \Cref{sec:sharedZ}, we establish causal
identification and estimate intervention distributions on this class
of general causal graphs.

\begin{figure*}
\centering
\begin{subfigure}[b]{0.66\textwidth}
\footnotesize
% \centering
% !TEX root = ../dcf_SEM.tex
  \begin{tabular}{lll}
    \toprule
    \textbf{Name} & \textbf{Children} & \textbf{Notation} \\
    \midrule
    Confounder & $\geq 1$ cause \& outcome & $U^{\mathrm{mlt}}$, $U^{\mathrm{sng}}$ \\
    \midrule
    Cause covariate & $\geq 1$ cause only & $W^{\mathrm{mlt}}$,  $W^{\mathrm{sng}}$ \\
    \midrule
    Outcome covariate & outcome only & $V$ \\
    \bottomrule
  \end{tabular}

%%% Local Variables:
%%% mode: latex
%%% TeX-master: "../dcf_SEM"
%%% End:

% \includegraphics[width=\textwidth]{img/glossary.pdf}
\caption{\label{fig:glossary}}
\end{subfigure}%
\begin{subfigure}[b]{0.33\textwidth}
\centering
\begin{adjustbox}{height=3.2cm}
% !TEX root = ../dcf_SEM.tex
\begin{tikzpicture}
% \tikzstyle{plate} = [draw, rectangle, rounded corners, fit=#1, dashed]
[obs/.style={draw,circle,filled,minimum size=1cm}, latent/.style={draw,circle,minimum size=1cm}]
  % Define nodes
  \node[latent] (u1mlt) {$U^\mathrm{mlt}_1$};
  \node[latent, right=0.5 of u1mlt] (u2mlt) {$U^\mathrm{mlt}_2$};
  \node[obs, right=1.2cm of u2mlt] (u1sng) {$U^\mathrm{sng}_1$};
  \node[latent, left=0.5 of u] (w1mlt) {$W^\mathrm{mlt}_1$};
  \node[latent, left=0.5 of w1mlt] (w1sng) {$W^\mathrm{sng}_1$};
  \node[obs, below=0.8cm of u, xshift=3cm] (am) {$A_m$};
  % \node[obs, below=0.6cm of u, xshift=2cm] (am1) {$A_{m-1}$};
  \node[obs, below=0.8cm of u, xshift=-3cm] (a1) {$A_1$};
  \node[obs, below=0.8cm of u, xshift=-1.5cm] (a2) {$A_2$};
  \node[obs, below=2.4cm of u] (y) {$Y$};
  \node[latent, below=0.8cm of y] (v) {$V$};
  \node[obs, accepting, below=0.8cm of y, xshift=3cm] (s) {$S$};
  % Connect the nodes
  \edge{w1sng} {a1};
  \edge{w1mlt} {a2, am};
  \edge{u1mlt} {a1, a2, y};
  \edge{u2mlt} {a2, am, y, s};
  \edge{u1sng} {am, y, s};
  % \edge{am} {s};
  \edge{v} {y};
  \edge{a1} {y};
  \edge{a2} {y};
  \edge{am} {y};
  \node at ($(a2)!.5!(am)$) {\ldots \quad \ldots};
\end{tikzpicture}
\end{adjustbox}
\caption{\label{fig:general}}
\end{subfigure}
\caption{(a) Types of nodes (b) The class of more general causal
  graphs. $S$ is the selection operator.}
\end{figure*}
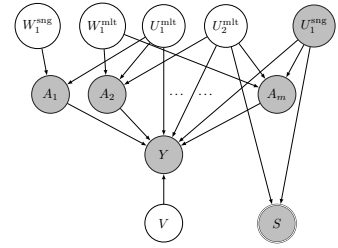

We focus on the class of general causal graphs in
\Cref{fig:general}.\footnote{There exist causal graphs that do not
  fall in this class; we leave them for future work.} As abvove, it
has $m$ causes $A_{1:m}$ and an outcome $Y$.  The goal is to estimate
$P(y \g \mathrm{do}(a_\mathcal{C}))$, where
$A_\mathcal{C}\subset \{A_1, \ldots, A_m\}$ is a subset of causes on
which we intervene. Apart from the causes and the outcome, the causal
graph has a few other types of variables. (\Cref{fig:glossary}
contains a glossary of terms.)

\parhead{Confounders.} Confounders are parents of \emph{both} the
causes and the outcome; they can be unobserved. In \Cref{fig:general},
for example, $U^{\mathrm{sng}}_i$ and $U^{\mathrm{mlt}}_i$ are
confounders; they have arrows into the outcome $Y$ and at least one of
the causes $A_i$. We differentiate between \emph{single-cause} and
\emph{multi-cause} confounders. Single-cause confounders like
$U^{\mathrm{sng}}_i$ affect only one cause; multi-cause confounders
like $U^{\mathrm{mlt}}_i$ affect two or more causes.

\parhead{Covariates.} There are two types of covariates---cause
covariates and outcome covariates.  Cause covariates are parents of
the causes, but \emph{not} the outcome; they can be unobserved. As
with confounders, we differentiate between \emph{single-cause}
covariates $W^{\mathrm{sng}}_i$ and \emph{multi-cause} covariates
$W^{\mathrm{mlt}}_i$.  Outcome covariates like $V$ are parents of the
outcome but \emph{not} the causes.  They do not affect any of the $m$
causes; they can be unobserved.

\parhead{Selection operator.} Following
\citet{bareinboim2012controlling}, we introduce a selection operator
$S \in \{0,1\}$ into the causal graph. The value $S = 1$ indicates an
individual being selected; otherwise, $S=0$. We only observe the
outcome of those individuals with $S = 1$, but we may observe the
causes on unselected individuals.  (E.g., consider a genome-wide
association study where we collect an expensive-to-measure trait on a
subset of the population but have genome data on a much larger set.)
Note that \Cref{fig:general} allows selection to occur on the
confounders.

These variables---confounders, covariates, and the selection
operator---compose the more general causal graphs with multiple
causes. We study identification and estimation on these causal graphs.

We extend the results around causal identification and estimation
under shared confounding (\Cref{thm:idsharedZ,thm:dcfcorrect-sharedZ})
to more general graphs. We first reduce the general graph
(\Cref{fig:general}) to one close to the shared confounding case; then
we handle the complications of selection bias. These steps lead to
causal identification and the correctness of the deconfounder on
general causal graphs.~\looseness=-1

To reduce the general graph of \Cref{fig:general}, we bundle all the
unobserved multi-cause confounders and null confounders
$\{U^{\mathrm{mlt}}, W^{\mathrm{mlt}}\}$ into a single unobserved
confounder $Z$. This variable $Z$ is shared by all the causes as in
\Cref{fig:sharedZ} and renders all the causes conditionally
independent. Moreover, it is sufficient to adjust for $Z$ and
single-cause confounders $U^{\mathrm{sng}}$ to estimate
$P(y\g \mathrm{do}(a_\mathcal{C}))$ because
$\{U^{\mathrm{mlt}}, W^{\mathrm{mlt}}, U^{\mathrm{sng}}\}$ constitute
an admissible set.

The general graph of \Cref{fig:general} also involves a selection
operator $S$; it allows an individual to be selected into the study
based on any unobserved multi-cause confounder $U^{\mathrm{mlt}}$ or
observed single-cause confounder $U^{\mathrm{sng}}$. We handle this
selection bias by assuming additional access to the
non-selection-biased distribution of the causes. This assumption
aligns with common conditions required by recovery under selection
bias (e.g., Theorem 2 of \citet{bareinboim2014recovering}).

We further extend the deconfounder algorithm to account for this
selection bias: we construct a substitute confounder based on this
non-selection-biased distribution of the causes; then we fit an
outcome model to the selection-biased joint distribution of the causes
and the outcome.  We show that the intervention distribution
$P(y\g\mathrm{do}(a_\mathcal{C}))$ is identifiable and the
deconfounder estimate $\hat{P}(y\g\mathrm{do}(a_\mathcal{C}))$ is
correct under assumptions analogous to those of
\Cref{thm:idsharedZ,thm:dcfcorrect-sharedZ}. The details of these
results are in \Cref{sec:causalidgensupp}.

%%% Local Variables:
%%% mode: latex
%%% TeX-master: "dcf_SEM"
%%% End:

% \input{sec_example}
% !TEX root = dcf_SEM.tex
\section{Discussion}
\label{sec:discussion}

We take a causal graphical view of \citet{wang2018blessings}.  By
treating some causes as proxies of the shared confounder, we can
identify the intervention distributions of the other causes. For a
general class of causal graphs, we prove that the intervention
distribution of subsets of causes is identifiable. We further show
that the deconfounder algorithm makes valid inferences of these
intervention distributions, a result that justifies the deconfounder
on causal graphs. The results of this paper generalize the theory in
\citet{wang2018blessings} and extend the applicability of the
deconfounder.

%%% Local Variables:
%%% mode: latex
%%% TeX-master: "dcf_SEM"
%%% End:

\appendix
% !TEX root = dcf_SEM.tex

\section{Proof of \Cref{thm:idsharedZ}}

\label{subsec:proof-idsharedZ}

\begin{proof} The proof of \Cref{thm:idsharedZ} relies on two
observations. The first observation starts with the integral equation
we solve:
\begin{align}
P(y\g a_\mathcal{C}, f(a_\mathcal{N}))
=&\int h(y, a_\mathcal{C},
a_\mathcal{X}) P(a_\mathcal{X}\g a_\mathcal{C}, f(a_\mathcal{N}))\dif a_\mathcal{X} \\
=&\int \int h(y, a_\mathcal{C},
a_\mathcal{X}) P(a_\mathcal{X}\g u) P(u \g a_\mathcal{C}, f(a_\mathcal{N}))\dif a_\mathcal{X}\dif u.\label{eq:obs1}
\end{align}

The first equality is due to \Cref{eq:integral}. The second equality
is due to the conditional independence implied by \Cref{fig:sharedZ}:
$A_\mathcal{X}\perp A_\mathcal{C}, f(a_\mathcal{N})\g U.$

The second observation relies on the null proxy:
\begin{align}
P(y\g a_\mathcal{C}, f(a_\mathcal{N})) 
=& \int P(y\g u, a_\mathcal{C}, f(a_\mathcal{N})) P(u\g a_\mathcal{C}, f(a_\mathcal{N}))\dif u\\
=& \int P(y\g u, a_\mathcal{C}) P(u \g a_\mathcal{C}, f(a_\mathcal{N}))\dif u.\label{eq:obs2}
\end{align}
The first equality is due to the definition of conditional
probability. The second equality is due to the second part of
\Cref{assumption:null-proxy-sharedZ}, which implies $Y\perp
f(a_\mathcal{N})\g U, A_\mathcal{C}.$ The reason is that
\begin{align}
P(y\g u, a_\mathcal{C}, f(a_\mathcal{N}))
=&\int P(y\g u, a_\mathcal{C}, a_\mathcal{X}, f(a_\mathcal{N})) P(a_\mathcal{X}\g u, a_\mathcal{C}, f(a_\mathcal{N}))\dif a_\mathcal{X}\\
=&\int P(y\g u, a_\mathcal{C}, a_\mathcal{X}) P(a_\mathcal{X}\g u, a_\mathcal{C})\dif a_\mathcal{X}\\
=& P(y\g u, a_\mathcal{C}).
\end{align}

In fact, it is sufficient to assume $Y\perp f(a_\mathcal{N})\g U,
A_\mathcal{C}$ instead of $Y\perp f(a_\mathcal{N})\g U, A_\mathcal{C},
A_\mathcal{X}$ in \Cref{thm:idsharedZ}. However, the latter is easier
to check and interpret.

Comparing \Cref{eq:obs1} and \Cref{eq:obs2} gives 
\begin{align}
\int \left[P(y\g u, a_\mathcal{C})-\int h(y, a_\mathcal{C},
a_\mathcal{X}) P(a_\mathcal{X}\g u)\dif a_\mathcal{X}\right]
 \times P(u\g a_\mathcal{C}, f(a_\mathcal{N}))\dif u = 0,
\end{align}
which, by the completeness condition in
\Cref{assumption:null-proxy-sharedZ}.2, implies
\begin{align}
\label{eq:fullcond}
P(y\g u, a_\mathcal{C}) = \int h(y, a_\mathcal{C},
a_\mathcal{X}) P(a_\mathcal{X}\g u)\dif a_\mathcal{X}.
\end{align}

\Cref{eq:fullcond} leads to identification:
\begin{align}
P(y\g \mathrm{do}(a_\mathcal{C})) 
=& \int \int h(y, a_\mathcal{C},
a_\mathcal{X}) P(a_\mathcal{X}\g u)\dif a_\mathcal{X} P(u)\dif u\\
=& \int h(y, a_\mathcal{C}, a_\mathcal{X}) P(a_\mathcal{X}) \dif a_\mathcal{X}.
\end{align}

Consider the special case of a single cause as in \Cref{fig:proxy}.
Let $a_\mathcal{C} = \{A_1\}$, $a_\mathcal{X} = \{X\}$,
$a_\mathcal{N}=N$, and $f(a_\mathcal{N}) = N$. The above proof reduces
to the identification proof for proxy variables (Theorem 1 of
\citet{miao2018identifying}). \end{proof}

%%% Local Variables:
%%% mode: latex
%%% TeX-master: "dcf_SEM"
%%% End:

\clearpage
\putbib[BIB1]
\end{bibunit}

\clearpage
\begin{bibunit}[alp]
{\onecolumn
% !TEX root = dcf_SEM.tex
\appendix
\onecolumn
{\Large\textbf{Appendix}}

\section{Causal identification and estimation on general causal graphs}

\label{sec:causalidgensupp}

In this section, we discuss causal identification and estimation on
general causal graphs in details.

\subsection{Causal identification on general causal graphs}

We first discuss the reduction to shared confounding and why it is
sufficient to adjust for $Z$ and single-cause confounders
$U^{\mathrm{sng}}$ to estimate $P(y\g
\mathrm{do}(a_\mathcal{C}))$.

The key to this reduction is the following observation: a shared
confounder $Z$ must ``capture''\footnote{The random variable $A$
``captures'' the random variable $B$ if $A$ contains all the
information of $B$. Technically, it means the sigma algebra of the
former is large than or equal to that of the latter:
$\sigma(B)\subset\sigma(A)$.} all multi-cause confounders
$U^{\mathrm{mlt}}$ and multi-cause null confounders $W^{\mathrm{mlt}}$
because $Z$ renders all the causes conditionally independent. 

\begin{figure}[h!]
\centering
\begin{adjustbox}{height=3.2cm}
% !TEX root = ../dcf_SEM.tex
\begin{tikzpicture}
% \tikzstyle{plate} = [draw, rectangle, rounded corners, fit=#1, dashed]
[obs/.style={draw,circle,filled,minimum size=1cm}, latent/.style={draw,circle,minimum size=1cm}]
  % Define nodes
  \node[latent] (Z) {$Z$};
  % \node[latent, right=0.5 of u1mlt] (u2mlt) {$U^\mathrm{mlt}_2$};
  \node[obs, right=3.0cm of Z] (u1sng) {$U^\mathrm{sng}_1$};
  % \node[latent, left=0.5 of u] (w1mlt) {$W^\mathrm{mlt}_1$};
  % \node[latent, left=0.5 of w1mlt] (w1sng) {$W^\mathrm{sng}_1$};
  \node[obs, below=0.8cm of u, xshift=3cm] (am) {$A_m$};
  % \node[obs, below=0.6cm of u, xshift=2cm] (am1) {$A_{m-1}$};
  \node[obs, below=0.8cm of u, xshift=-3cm] (a1) {$A_1$};
  \node[obs, below=0.8cm of u, xshift=-1.5cm] (a2) {$A_2$};
  \node[obs, below=2.4cm of u] (y) {$Y$};
  \node[latent, below=0.8cm of y] (v) {$V$};
  \node[obs, accepting, below=0.8cm of y, xshift=3cm] (s) {$S$};
  % Connect the nodes
  % \edge{w1sng} {a1};
  % \edge{w1mlt} {a2, am};
  \edge{Z} {a1, a2, am, y, s};
  % \edge{u2mlt} {a2, am, y, s};
  \edge{u1sng} {am, y, s};
  \edge{am} {s};
  \edge{v} {y};
  \edge{a1} {y};
  \edge{a2} {y};
  \edge{am} {y};
  \node at ($(a2)!.5!(am)$) {\ldots \quad \ldots};
\end{tikzpicture}
\end{adjustbox}
\caption{\label{fig:generalredZ} The reduced causal graph with shared
confounding. }
\end{figure}
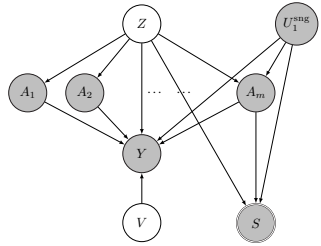%

More concretely, consider a random variable $Z$ that renders all the
causes $A_1,
\ldots, A_m$ conditionally independent as in \Cref{fig:generalredZ}.
We claim that $Z$ must capture all the multi-cause confounders and
null confounders $\{U^{\mathrm{mlt}}, W^{\mathrm{mlt}}\}$. We can
prove this claim by contradiction. Imagine there exists some
multi-cause confounder $U^{\mathrm{mlt}}_i$ that is not captured by
$Z$. This multi-cause confounder $U^{\mathrm{mlt}}_i$ will induce
dependence among the causes because $U^{\mathrm{mlt}}_i$ affects two
or more causes by definition. Due to this dependence, the $m$ causes
could not have been conditionally independent given $Z$ because $Z$
does not capture $U^{\mathrm{mlt}}_i$. It contradicts the fact that
$Z$ renders all the causes conditionally independent. This argument
shares the same spirit with the substitute confounder argument in
\citet{wang2018blessings}.

Further, assume the general causal graph \Cref{fig:general}. Recall
that the goal is to estimate the intervention distribution of the
causes $P(y\g
\mathrm{do}(a_\mathcal{C}))$. In this graph, the set of single-cause
confounders $U^{\mathrm{sng}}$, multi-cause confounders
$U^{\mathrm{mlt}}$, and multi-cause null confounders
$W^{\mathrm{mlt}}$ constitute an admissible set; the set of random
variables $\{W^{\mathrm{mlt}}, U^{\mathrm{mlt}}, U^{\mathrm{sng}}\}$
block all back-door paths. While it is unnecessary to adjust for the
null confounders $W^{\mathrm{mlt}}$, adjusting for $W^{\mathrm{mlt}}$
is still valid: it does not open any back-door paths. We note that
this set $\{W^{\mathrm{mlt}}, U^{\mathrm{mlt}}, U^{\mathrm{sng}}\}$,
while admissible, might not be observed.

Following this discussion, we can reduce all the multi-cause
confounders and null confounders $\{U^{\mathrm{mlt}},
W^{\mathrm{mlt}}\}$ into a shared confounder $Z$. Therefore, assuming
the general causal graph (\Cref{fig:general}), we can equivalently
identify the intervention distributions $P(y\g
\mathrm{do}(a_\mathcal{C}))$ using a reduced causal graph
(\Cref{fig:generalredZ}); it involves only the single-cause
confounders $U^{\mathrm{sng}}$ and a shared confounder $Z$.

Together with the discussion in \Cref{sec:general-intervention-id}, we
formally state the validity of the reduction from general causal
graphs to ones with shared confounding.

\begin{lemma} (Validity of reduction)
\label{lemma:reduce} Assume the causal graph in \Cref{fig:general}.
Adjusting for the multi-cause confounders and null confounders on the
general causal graph \Cref{fig:general} is equivalent to adjusting for
the shared confounder in \Cref{fig:generalredZ}:
% \vspace{-6pt}
\begin{align}
\label{eq:reductionvalid}
P(y\g u^{\mathrm{sng}}, u^{\mathrm{mlt}}, w^{\mathrm{mlt}}, a_1,
\ldots, a_m, s = 1)
=P(y\g u^{\mathrm{sng}}, z, a_1,
\ldots, a_m, s = 1).
\end{align}
\end{lemma}

\begin{proofsk} The proof uses a measure-theoretic argument to
characterize the information contained in the $Z$ variable in
\Cref{fig:generalredZ}. Roughly, the information in $Z$ is same as the
information of all multi-cause confounders, all null confounders, and
some independent error:
\begin{align}
\label{eq:z-info}
\sigma(z) = \sigma(u^{\mathrm{mlt}}, w^{\mathrm{mlt}}, \epsilon_Z),
\end{align}
where $\sigma(\cdot)$ denotes the $\sigma$-algebra of a random
variable. The independent error $\epsilon_Z$ satisfies
\[\epsilon_Z\perp Y, S, U^{\mathrm{sng}}, U^{\mathrm{mlt}},
W^{\mathrm{mlt}}, A_1, \ldots, A_m.\] 
\Cref{eq:z-info} implies that
conditioning on $Z$ is equivalent to conditioning on
$U^{\mathrm{mlt}}, W^{\mathrm{mlt}}, \epsilon_Z$; it leads to
\Cref{eq:reductionvalid}. The full proof is in
\Cref{sec:reduce-proof}.
\end{proofsk}

\parhead{Causal identification on the reduced causal graph
(\Cref{fig:generalredZ}).} We have just reduced general causal graphs
(\Cref{fig:general}) to one with shared confounding
(\Cref{fig:generalredZ}). This reduction allows us to establish causal
identification on general causal graphs. We extend
\Cref{thm:idsharedZ} from \Cref{fig:sharedZ} to
\Cref{fig:generalredZ}. With the reduction step (\Cref{lemma:reduce}),
it leads to causal identification on general causal graphs.

How can we identify the intervention distributions $P(y\g
\mathrm{do}(a_\mathcal{C}))$ on the reduced graph
(\Cref{fig:generalredZ})? \Cref{fig:generalredZ} has a confounder $Z$
that is shared across all causes. This structure is similar to the
unobserved shared confounding \Cref{fig:sharedZ}. In addition to the
shared confounder $Z$, the reduced graph involves single-cause
confounders $U^{\mathrm{sng}}$ and the selection operator $S$. We
posit two assumptions on them to enable causal identification.

\begin{assumption} 
\label{assumption:generalredZ}
The causal graph \Cref{fig:generalredZ} satisfies the following
conditions:
\begin{enumerate}
	\item All single-cause confounders $U^{\mathrm{sng}}_i$'s are observed.
	\item The selection operator $S$ satisfies
	\begin{align}
	S\perp (A, Y)\g  Z, U^{\mathrm{sng}}.
	\end{align}
	\item We observe the non-selection-biased distribution \[P(a_1, \ldots, a_m,
u^{\mathrm{sng}})\] and the selection-biased distribution \[P(y,
u^{\mathrm{sng}}, a_1, \ldots, a_m \g s=1).\]
\end{enumerate}
\end{assumption}

\Cref{assumption:generalredZ}.1 requires that the confounders that
affect the outcome and only one of the causes must be observed. It
allows us to adjust for confounding due to these single-cause
confounders. \Cref{assumption:generalredZ}.2 roughly requires that
selection can only occur on the confounders.
\Cref{assumption:generalredZ}.3 requires access to the
non-selection-biased distribution of the causes and
single-cause-confounders. It aligns with common conditions required by
recovery under selection bias (e.g., Theorem 2 of
\citet{bareinboim2014recovering}).

We next establish causal identification on the reduced causal graph
\Cref{fig:generalredZ}. We additionally make
\Cref{assumption:null-proxy-generalredZ}; it is a variant of
\Cref{assumption:null-proxy-sharedZ} but involves single-cause
confounders and the selection operator.

\begin{assumption}
\label{assumption:null-proxy-generalredZ}
There exists some function $f$ and a set
$\emptyset \ne \mathcal{N} \subset \{1, \ldots, m\}\backslash\mathcal{C}$
such that
% \vspace{-6pt}
\begin{enumerate}
\item The outcome $Y$ does not causally depend on $f(A_\mathcal{N})$:
\begin{align} 
\label{eq:nullproxyreqgeneral}
f(A_\mathcal{N})\perp Y \g Z, A_\mathcal{C}, A_\mathcal{X}, U^{\mathrm{sng}}, S = 1
\end{align}
where $\mathcal{X}=  \{1, \ldots, m\}\backslash
(\mathcal{C}\cup \mathcal{N}) \ne \emptyset.$
\item The conditional $P(z\g a_\mathcal{C},
f(a_\mathcal{N}), u^{\mathrm{sng}}_\mathcal{C}, s=1)$ is complete in
$f(a_\mathcal{N})$  for almost all $a_\mathcal{C}$ and
$u^{\mathrm{sng}}_\mathcal{C}$, where $U^{\mathrm{sng}}_\mathcal{C}$
is the single-cause confounders affecting $A_\mathcal{C}$.
\item The conditional $P(f(a_\mathcal{N})\g
a_\mathcal{C}, a_\mathcal{X}, u^{\mathrm{sng}}_\mathcal{C}, s=1)$ is complete in $a_\mathcal{X}$ for
almost all $a_\mathcal{C}$ and $u^{\mathrm{sng}}_\mathcal{C}$.
\end{enumerate}
\end{assumption}

Under \Cref{assumption:generalredZ} and
\Cref{assumption:null-proxy-generalredZ}, we can identify the
intervention distributions $P(y \g \mathrm{do}(a_\mathcal{C}))$.
\begin{lemma}
\label{lemma:idgeneralredZ}
Assume the causal graph \Cref{fig:generalredZ}. Under
\Cref{assumption:generalredZ} and
\Cref{assumption:null-proxy-generalredZ}, the intervention
distribution of the causes $A_\mathcal{C}$ is identifiable:
\begin{align}
\label{eq:general-adjust}
&P(y\g \mathrm{do}(a_\mathcal{C})) \\
=&\int \int h(y, a_\mathcal{C},
a_\mathcal{X}, u^{\mathrm{sng}}_{\mathcal{C}}) P(a_\mathcal{X}) P(u^{\mathrm{sng}}_{\mathcal{C}})\dif a_\mathcal{X} \dif u^{\mathrm{sng}}_{\mathcal{C}}\nonumber
\end{align}
for any solution $h$ to the integral equation
\begin{align}
\label{eq:integral-general}
&P(y\g a_\mathcal{C}, f(a_\mathcal{N}), u^{\mathrm{sng}}_{\mathcal{C}}, s=1)\int h(y, a_\mathcal{C},
a_\mathcal{X}, u^{\mathrm{sng}}_{\mathcal{C}})\times P(a_\mathcal{X}\g a_\mathcal{C}, f(a_\mathcal{N}), u^{\mathrm{sng}}_{\mathcal{C}}, s=1)\dif a_\mathcal{X},
\end{align}
where $U^{\mathrm{sng}}_\mathcal{C}$
is the single-cause confounders affecting $A_\mathcal{C}$.
Moreover, the solution to \Cref{eq:integral-general} always exists
under weak regularity conditions in \Cref{sec:integral-sol-exist}.
\end{lemma}

\begin{proofsk} The proof adopts a similar argument as in the proof of
\Cref{thm:idsharedZ}. We only need to take care of the additional
(observed) single-cause confounders and the selection operator. In
particular, \Cref{assumption:generalredZ}.2 lets us shift from the
selection biased distribution $P(y\g z, a_\mathcal{C},
u^{\mathrm{sng}}_{\mathcal{C}}, s=1)$ to the non-selection-biased one
$P(y\g z, a_\mathcal{C}, u^{\mathrm{sng}}_{\mathcal{C}})$.
The full proof is in \Cref{sec:id-generalredZ-proof}.
\end{proofsk}

\parhead{Causal identification on general causal graphs
(\Cref{fig:general}).} Based on the previous analysis on the reduced
graph, we establish causal identification result on general causal
graphs.

\begin{thm} 
\label{thm:idgeneral}
Assume the causal graph \Cref{fig:general}. Assume a variant of
\Cref{assumption:generalredZ} and
\Cref{assumption:null-proxy-generalredZ} (detailed in
\Cref{sec:id-general-proof}), the intervention distribution of the
causes $A_\mathcal{C}$ is identifiable using \Cref{eq:general-adjust}
and \Cref{eq:integral-general}.
\end{thm}
\begin{proofsk} This result is a direct consequence of
\Cref{lemma:reduce} and \Cref{lemma:idgeneralredZ}. The full proof is
in \Cref{sec:id-general-proof}.
\end{proofsk}

\subsection{Causal estimation with the deconfounder}
We finally prove the correctness of the deconfounder algorithm on
general causal graphs. We build on the identification result on
general causal graphs (\Cref{thm:idgeneral}). We then show that the
deconfounder provides correct causal estimates by implicitly solving
the integral equation (\Cref{eq:integral-general}). This argument is
similar to the argument of \Cref{thm:dcfcorrect-sharedZ}.

The deconfounder algorithm for general causal graphs with selection
bias extends the version described in \Cref{subsec:dcf-sharedZ}.
Specifically, \Cref{assumption:dcf-sharedZ} allows the deconfounder
algorithm to have access to both the non-selection-biased data $P(a_1,
\ldots, a_m, u^{\mathrm{sng}})$ and the selection-biased data $P(y,
u^{\mathrm{sng}}, a_1, \ldots, a_m \g s=1).$ In this case, the
deconfounder algorithm outputs two estimates:
\begin{align}
\label{eq:dcfgeneralreq}
\hat{P}(a_1, \ldots, a_m, u^{\mathrm{sng}}, \hat{z}) = 
\hat{P}(\hat{z})\hat{P}(u^{\mathrm{sng}}\g a_1, \ldots, a_m, \hat{z})\prod_{i=1}^n \hat{P}(a_i\g \hat{z}),
\end{align}
and
\begin{align*}
\hat{P}(y, a_1, \ldots, a_m, u^{\mathrm{sng}}, \hat{z}\g s = 1).
\end{align*}
We note that the former is constructed using only the causes $A_1,
\ldots, A_m$ and single-cause confounders $U^{\mathrm{sng}}$. Moreover,
both deconfounder estimates must be consistent with the observed data:
\begin{align*}
\int \hat{P}(a_1, \ldots, a_m, u^{\mathrm{sng}}, \hat{z})\dif \hat{z}
=
P(a_1, \ldots, a_m, u^{\mathrm{sng}}),
\end{align*}
\begin{align*}
\int \hat{P}(y, a_1, \ldots, a_m, u^{\mathrm{sng}}, \hat{z}\g s = 1)\dif \hat{z} 
= P(y, a_1, \ldots, a_m, u^{\mathrm{sng}}\g s=1).
\end{align*}

We note that the substitute confounder $\hat{Z}$ does not necessarily
coincide with the true confounders $U^{\mathrm{mlt}}$ or the true null
confounders $W^{\mathrm{mlt}}$. Nor do $\hat{P}(a_1, \ldots, a_m,
u^{\mathrm{sng}}, \hat{z})$ and $\hat{P}(y, a_1, \ldots, a_m,
u^{\mathrm{sng}}, \hat{z}\g s = 1)$ need to be unique. We will show
that any $\hat{Z}$ and $\hat{P}$ that the deconfounder outputs will
lead to a correct estimate of  $\hat{P}(y\g
\mathrm{do}(a_\mathcal{C}))$.

Finally the deconfounder estimates
\begin{align}
\label{eq:dcfgeneral}
&\hat{P}(y\g \mathrm{do}(a_\mathcal{C})) \stackrel{\Delta}{=}&\int \hat{P}(y\g a_1, \ldots, a_m,
\hat{z}, u^{\mathrm{sng}}_\mathcal{C}, s = 1) \times \hat{P}(a_{\{1, \ldots, m\}\backslash\mathcal{C}} , \hat{z})P(u^{\mathrm{sng}}_\mathcal{C})\dif u^{\mathrm{sng}}_\mathcal{C}\dif \hat{z}\dif a_{\{1, \ldots, m\}\backslash\mathcal{C}},
\end{align}
where $U^{\mathrm{sng}}_\mathcal{C}$ are the single-cause confounders
that affect the causes $A_\mathcal{C}$.

We now prove the correctness of the deconfounder on general causal
graphs. We make a variant of \Cref{assumption:dcf-sharedZ} and state
the correctness result.
\begin{assumption}
\label{assumption:dcf-generalZ}
The deconfounder outputs the estimates $\hat{P}(y, a_1, \ldots, a_m,
u^{\mathrm{sng}},
\hat{z}\g s = 1)$~and $\hat{P}(a_1, \ldots, a_m, u^{\mathrm{sng}},
\hat{z})$ that satisfy the following conditions:
\begin{enumerate}
\item It is consistent with \Cref{assumption:generalredZ}.1:
\begin{align}
\label{eq:dcfgeneral0}
\hat{P}(a_1, \ldots, a_m\g \hat{z}, u^{\mathrm{sng}}, s=1) \\
=\hat{P}(a_1, \ldots, a_m\g \hat{z}, u^{\mathrm{sng}}).
\end{align}
\item It is consistent with \Cref{assumption:null-proxy-generalredZ}.1:
\begin{align}
\label{eq:dcfgeneral1}
\hat{P}(y\g a_\mathcal{C}, a_\mathcal{X}, f(a_\mathcal{N}), 
\hat{z}, u^{\mathrm{sng}}, s = 1)\\ = \hat{P}(y\g a_\mathcal{C}, a_\mathcal{X}, 
\hat{z}, u^{\mathrm{sng}}, s = 1).
\end{align}
\item The conditional $\hat{P}(\hat{z}\g a_\mathcal{C},
a_\mathcal{X}, u^{\mathrm{sng}}, s=1)$ is complete in $a_\mathcal{X}$ for almost all
$a_\mathcal{C}$.
\end{enumerate}
The conditional $\hat{P}(\hat{z}\g a_\mathcal{C}, a_\mathcal{X},
u^{\mathrm{sng}}, s=1)$, \Cref{eq:dcfgeneral0}, and
\Cref{eq:dcfgeneral1} can all be computed from the deconfounder
estimate $\hat{P}(a_1,
\ldots, a_m, u^{\mathrm{sng}}, \hat{z})$ and $\hat{P}(y, a_1, \ldots,
a_m, u^{\mathrm{sng}}, \hat{z}\g s = 1)$.
\end{assumption}

\begin{thm} (Correctness of the deconfounder on general causal graphs)
\label{thm:dcfcorrect-general} 
Assume the causal graph \Cref{fig:general}. Assume a variant of
\Cref{assumption:generalredZ} and
\Cref{assumption:null-proxy-generalredZ} (detailed in
\Cref{sec:dcfcorrect-general-proof}). Under
\Cref{assumption:dcf-generalZ} and weak regularity conditions, the
deconfounder provides correct estimates of the intervention
distribution:
\begin{align}
\hat{P}(y\g \mathrm{do}(a_\mathcal{C})) = P(y\g \mathrm{do}(a_\mathcal{C})).
\end{align}
\end{thm}
% \vspace{-10pt}
\begin{proofsk} The proof of \Cref{thm:dcfcorrect-general} follows a
similar argument as in the proof of \Cref{thm:dcfcorrect-sharedZ}. We
only need to additionally take care of the single-cause confounders
and the selection operator. The full proof is in
\Cref{sec:dcfcorrect-general-proof}.
\end{proofsk}
% \vspace{-10pt}
\Cref{thm:dcfcorrect-general} establishes the correctness of the
deconfounder on general causal graphs under certain types of selection
bias. It justifies the deconfounder on general causal graphs.

\section{Examples of \Cref{assumption:null-proxy-sharedZ}}

\label{sec:example-null-proxy}

As an example, if the structural equation writes
\[Y = g(A_1+A_2, A_3, \ldots, A_m, U, \epsilon), \]
where $\epsilon\perp U, A_1, \ldots, A_m$, then
\Cref{assumption:null-proxy-sharedZ}.1 is satisfied if $A_1$ and $A_2$
are identically Gaussian: $A_\mathcal{N} = (A_1, A_2)$ and
$f(A_\mathcal{N}) = A_1-A_2$ satisfies
\[A_1-A_2\perp Y\g U, A_3, \ldots, A_m.\]

If $A_1$ and $A_2$ are both Gaussian but not identically distributed,
then $f(A_\mathcal{N}) = \alpha_1 A_1-\alpha_2 A_2$ would satisfy
\[\alpha_1 A_1-\alpha_2 A_2\perp Y\g U, A_3, \ldots, A_m,\]
for some constant $\alpha_1$ and $\alpha_2$.

Similarly, if the structural equation writes
\[Y = g(A_1\times A_2, A_3, \ldots, A_m, U, \epsilon), \]
where $\epsilon\perp U, A_1, \ldots, A_m$, then
\Cref{assumption:null-proxy-sharedZ}.1 is satisfied if $A_1$ and $A_2$
are identically log-normal: $A_\mathcal{N} = (A_1, A_2)$ and
$f(A_\mathcal{N}) = A_1 / A_2$ satisfies
\[A_1 / A_2\perp Y\g U, A_3, \ldots, A_m.\]

As a final example, if the structural equation writes
\[Y = g(A_1 \&\& A_2, A_3, \ldots, A_m, U, \epsilon), \]
where $\epsilon\perp U, A_1, \ldots, A_m$ and $A_1, A_2$ are both
binary, then \Cref{assumption:null-proxy-sharedZ}.1 is satisfied:
$A_\mathcal{N} = (A_1, A_2)$ and $f(A_\mathcal{N}) = A_1 \textrm{ XOR }
A_2$ satisfies
\[A_1 \textrm{ XOR }
A_2\perp Y\g U, A_3, \ldots, A_m.\]

% !TEX root = dcf_SEM.tex

\section{Example: A linear causal model} 

\label{subsec:examplecontd}

We illustrate \Cref{thm:idgeneral,thm:dcfcorrect-general} in a linear
causal model.

Consider the meal/body-fat example. The causes are ten types of food
$A_1, \ldots, A_{10}$; the outcome is a person's body fat $Y$.  How
does food consumption affect body fat?

In this example, the individual's lifestyle $U^\mathrm{mlt}$ is a
multi-cause confounder. Whether a person is vegan $W^\mathrm{mlt}$ is
a multi-cause null confounder. Both $U^\mathrm{mlt}$ and
$W^\mathrm{mlt}$ are unobserved. Whether one has easy access to good
burger shops $U^\mathrm{sng}$ is a single-cause confounder; it affects
both burger consumption $A_1$ and body fat percentage $Y$;
$U^\mathrm{sng}$ is observed. Finally, the observational data comes
from a survey with selection bias $S$; people with healthy lifestyle
are more likely to complete the survey.

Every variable is associated with a disturbance term $\epsilon$, which
comes from a standard normal.  Given these variables, suppose the real
world is linear,
\begin{align*}
  &U^\mathrm{mlt} = \epsilon_{U^\mathrm{mlt}},
    U^\mathrm{sng} = \epsilon_{U^\mathrm{sng}},
    W^\mathrm{mlt} = \epsilon_{W^\mathrm{mlt}},\\
  &A_1 = \alpha_{A_1U}U^\mathrm{mlt} + \alpha_{A_1W}W^\mathrm{mlt} + \alpha_{A_1U'}U^\mathrm{sng} +\epsilon_{A_1},\\
  &A_i = \alpha_{A_iU}U^\mathrm{mlt} + \alpha_{A_iW}W^\mathrm{mlt} + \epsilon_{A_i}, i=2, \ldots, 10,\\
  &Y = \sum_{i=1}^{10}\alpha_{YA_i} A_i + \alpha_{YU}U^\mathrm{mlt} + \alpha_{YU'}U^\mathrm{sng}+\epsilon_{Y}.
\end{align*}
These equations describe the true causal model of the world. The
confounders and null confounders $\{U^\mathrm{mlt}, W^\mathrm{mlt}\}$
are unobserved.

We are interested in the intervention distribution of the first two
food categories, burger $(A_1)$ and broccoli $(A_2)$:
$P(y\g \mathrm{do}(a_1, a_2))$. (We emphasize that we might be
interested in any subsets of the causes.)  This world satisfies the
assumptions of \Cref{thm:idgeneral}. Even though the confounders
$U^\mathrm{mlt}$ are unobserved, the intervention distribution
$P(y\g \mathrm{do}(a_1, a_2))$ is identifiable.

Now consider a simple deconfounder. Fit a 2-D \gls{PPCA} to the data
about food consumption $\{A_1, \ldots, A_{10}\}$; we do not model the
outcome $Y$.  \citet{wang2018blessings} also checks the model to
ensure it fits the distribution of the assigned causes.  (Let's assume
that 2-D \gls{PPCA} passes this check.)

\gls{PPCA} leads to a linear estimate of the substitute confounder,
\begin{align}
  \hat{Z} = \left(\sum_{i=1}^{10}\gamma_{1i} A_i + \epsilon_{1\hat{Z}},
  \sum_{i=1}^{10}\gamma_{2i} A_i + \epsilon_{2\hat{Z}}\right),
\end{align}
for parameters $\gamma_{1i}$ and $\gamma_{2i}$, and Gaussian noise
$\epsilon_{i,\hat{Z}}$.

This substitute confounder $\hat{Z}$ satisfies
\Cref{assumption:dcf-generalZ}. Plausibly, the real world satisfies
the variant of \Cref{assumption:generalredZ} and
\Cref{assumption:null-proxy-generalredZ}. These assumptions greenlight
us to calculate the intervention distribution. We fit an outcome model
using the substitute confounder $\hat{Z}$ and calculate the
intervention distribution using \Cref{eq:dcfgeneral}.
\Cref{thm:dcfcorrect-general} guarantees that this estimate is
correct.

%%% Local Variables:
%%% mode: latex
%%% TeX-master: "dcf_SEM"
%%% End:

\section{Proof of \Cref{thm:dcfcorrect-sharedZ}}

\label{subsec:proof-dcfcorrect}

\begin{proof} 

\Cref{assumption:dcf-sharedZ}.2 guarantees the existence of some
function $\hat{h}$ such that
\begin{align}
\label{eq:dcf2supp}
\hat{P}(y\g a_\mathcal{C},
\hat{z}) = \int \hat{h}(y, a_\mathcal{C},
a_\mathcal{X}) \hat{P}(a_\mathcal{X}\g \hat{z})\dif a_\mathcal{X}
\end{align}
under weak regularity conditions. (We will discuss the reason in
\Cref{sec:integral-sol-exist}.)

We first claim that $\hat{h}(y, a_\mathcal{C}, a_\mathcal{X})$ solves
\begin{align}
\label{eq:keyeq}
P(y\g a_\mathcal{C}, f(a_\mathcal{N})) = \int \hat{h}(y,
a_\mathcal{C},
a_\mathcal{X}) P(a_\mathcal{X}\g a_\mathcal{C}, f(a_\mathcal{N}))\dif a_\mathcal{X}.
\end{align}

Given this claim (\Cref{eq:keyeq}), we have
\begin{align*}
&\hat{P}(y\g \mathrm{do}(a_\mathcal{C})) \\
=& \int \hat{P}(y\g \hat{z},
a_\mathcal{C})\hat{P}(\hat{z})\dif \hat{z}\\
=& \int \hat{h}(y, a_\mathcal{C},
a_\mathcal{X}) \hat{P}(a_\mathcal{X}\g \hat{z})\dif a_\mathcal{X}\hat{P}(\hat{z})\dif \hat{z}\\
=& \int \hat{h}(y, a_\mathcal{C},
a_\mathcal{X}) P(a_\mathcal{X})\dif a_\mathcal{X}\\
=& P(y\g \mathrm{do}(a_\mathcal{C})),
\end{align*}
which proves the theorem. The first equality is due to \Cref{eq:dcf};
the second is due to \Cref{eq:keyeq}; the third is due to the
deconfounder estimate being consistent with the observed data
distribution by construction; the fourth is due to the above claim
(\Cref{eq:keyeq}) and \Cref{thm:idsharedZ}.

We next prove the claim (\Cref{eq:keyeq}). Start with the right side of
the equality.
\begin{align*}
&\int \hat{h}(y,
a_\mathcal{C},
a_\mathcal{X}) P(a_\mathcal{X}\g a_\mathcal{C}, f(a_\mathcal{N}))\dif a_\mathcal{X}\\
=& \int \int \hat{h}(y,
a_\mathcal{C},
a_\mathcal{X}) \hat{P}(a_\mathcal{X}\g \hat{z}) \hat{P}(\hat{z}\g a_\mathcal{C}, f(a_\mathcal{N}))\dif a_\mathcal{X} \dif \hat{z}\\
=& \int \hat{P}(y\g a_\mathcal{C}, \hat{z}) \hat{P}(\hat{z}\g a_\mathcal{C}, f(a_\mathcal{N}))\dif \hat{z}\\
=& P(y\g a_\mathcal{C}, f(a_\mathcal{N})),
\end{align*}

which establishes the claim. The first equality is due to
\Cref{eq:dcfreq1} and the deconfounder estimate being consistent with
the observed data; the second is due to \Cref{eq:dcf2supp}; the third
is due to \Cref{assumption:dcf-sharedZ}.1, which implies
\begin{align} 
\label{eq:dcf1supp}
\hat{P}(y\g a_\mathcal{C}, f(a_\mathcal{N}), 
\hat{z}) = \hat{P}(y\g a_\mathcal{C},
\hat{z}).
\end{align}

Similar to \Cref{assumption:null-proxy-sharedZ}.1, it is sufficient to
assume \Cref{eq:dcf1supp} directly. However,
\Cref{assumption:dcf-sharedZ}.1 is easier to check and more
interpretable; it directly relates to the deconfounder outcome model.

\end{proof}

\section{Existence of solutions to the integral equations }
\label{sec:integral-sol-exist}

\Cref{thm:idsharedZ} involves solving the integral equation 
\begin{align}
\label{eq:integral-supp}
P(y\g a_\mathcal{C}, f(a_\mathcal{N})) = \int h(y, a_\mathcal{C},
a_\mathcal{X}) P(a_\mathcal{X}\g a_\mathcal{C}, f(a_\mathcal{N}))\dif
a_\mathcal{X}.
\end{align}

When does a solution exist for \Cref{eq:integral-supp}? We appeal to
Proposition 1 of \citet{miao2018identifying}.

\begin{prop}(Proposition 1 of \citet{miao2018identifying}) 
\label{prop:solexistprop}
Denote
$L^2\{F(t)\}$ as the space of all square-integrable function of $t$
with respect to a c.d.f. $F(t)$. A solution to integral equation
\begin{align}
P(y\g z, x) = \int h(w,x,y)P(w\g z,x)\dif w
\end{align}
exists if 
\begin{enumerate}
	\item the conditional distribution $P(z\g w,x)$ is complete in $w$ for all $x$,
	\item $\int \int P(w\g z, x) P(z\g w, x)\dif w \dif z < +\infty$,
	\item $\int [P(y\g z, x)]^2 P(z\g x)\dif z < + \infty,$
	\item $\sum_{n=1}^{+\infty} |<P(y\g z, x), \psi_{x, n}>|^2 < +\infty,$
\end{enumerate}
where the inner product is $<g, h> = \int g(t)h(t)\dif F(t)$, and
$(\lambda_{x,n}, \phi_{x,n}, \psi_{x,n})_{n=1}^\infty$ is a singular
value decomposition of the conditional expectation operator $K_x:
L^2\{F(w\g x)\}\rightarrow L^2\{F(z\g x)\}, K_x(h) = \E{}{h(w)\g z,
x}$ for $h\in L^2\{F(w\g x)\}.$
\end{prop}

Leveraging \Cref{prop:solexistprop}, we can establish sufficient
conditions for existence of a solution to \Cref{eq:integral-supp}.
\begin{corollary}
A solution exist for the integral equation \Cref{eq:integral-supp} if
\begin{enumerate}
	\item the conditional distribution $P(f(a_\mathcal{N})\g a_\mathcal{X},a_\mathcal{C})$ is complete in $a_\mathcal{X}$ for all $a_\mathcal{C}$,
	\item $\int \int P(a_\mathcal{X}\g f(a_\mathcal{N}), a_\mathcal{C}) P(f(a_\mathcal{N}) \g a_\mathcal{X}, a_\mathcal{C})\dif a_\mathcal{X} \dif f(a_\mathcal{N}) < +\infty$,
	\item $\int [P(y\g f(a_\mathcal{N}), a_\mathcal{C})]^2 P(f(a_\mathcal{N})\g a_\mathcal{C})\dif f(a_\mathcal{N}) < + \infty,$
	\item $\sum_{n=1}^{+\infty} |<P(y\g f(a_\mathcal{N}), a_\mathcal{C}), \psi_{a_\mathcal{C}, n}>|^2 < +\infty,$
\end{enumerate}
where $\psi_{a_\mathcal{C}, n}$ is similarly defined as a component of
the singular value decomposition.
\end{corollary}

We remark that the first condition is precisely
\Cref{thm:idsharedZ}.3; others are weak regularity conditions.

By the same token, we can establish sufficient conditions for solution existence of \Cref{eq:dcf2}, \Cref{eq:integral-general}. The same argument also applies to the integral equation involved in \Cref{thm:dcfcorrect-general}:
\begin{align}
\hat{P}(y\g a_\mathcal{C},
\hat{z}, u^{\mathrm{sng}}_\mathcal{C}, s = 1) =\int \hat{h}(y, a_\mathcal{C},
a_\mathcal{X}, u^{\mathrm{sng}}_\mathcal{C}) \hat{P}(a_\mathcal{X}\g \hat{z}, u^{\mathrm{sng}}_\mathcal{C}, s = 1)\dif a_\mathcal{X}.
\end{align}
It is easy to show that the conditions described in the main text are
sufficient to guarantee the existence of solutions under weak
regularity conditions. We omit the details here.

\section{Proof of \Cref{lemma:reduce}}

\label{sec:reduce-proof}

The idea of the proof is to start with the structural equations of the
general causal graph \Cref{fig:general}. Then posit the existence of a
latent variable $Z$ that renders all the causes conditionally
independent; \Cref{fig:generalredZ} features this conditional
independence structure. We will quantify the information (i.e. the
$\sigma$-algebra) of this latent variable $Z$; $Z$ contains the
information of the union of multi-cause confounders
$U^{\mathrm{mlt}}$, multi-cause null confounders $W^{\mathrm{mlt}}$,
and some independent error. This result lets us establish
\begin{align}
P(y\g u^{\mathrm{sng}}, u^{\mathrm{mlt}}, w^{\mathrm{mlt}}, a_1,
\ldots, a_m, s = 1)
=P(y\g u^{\mathrm{sng}}, z, a_1,
\ldots, a_m, s = 1).
\end{align}

We start with a generic structural equation model for multiple causes.
\begin{align} 
W_k &= f_{W_k}(\epsilon_{W_k}), &k= 1,\ldots, K,
K\geq0,\label{eq:conf}\\
U_j &= f_{U_j}(\epsilon_{U_j}), &j= 1,\ldots, J,
J\geq0,\label{eq:nullconf}\\
V_l &= f_{V_l}(\epsilon_{V_l}), &l= 1,\ldots, L,
L\geq0,\label{eq:covar}\\
A_{i} &= f_{A_i}(W_{S^W_{A_i}}, U_{S^U_{A_i}}, \epsilon_{A_i}),&
i=1,\ldots,m, m\geq2,\label{eq:cause}\\
Y &= f_y(A_1,\ldots, A_m, U_1, \ldots, U_K, V_1,\ldots V_L,
\epsilon_{Y})\label{eq:outcome},
\end{align} 
where all the errors $\epsilon_{W_k}, \epsilon_{U_j},
\epsilon_{V_l},
\epsilon_{A_i}, \epsilon_{Y}$ are independent. Notation wise, we note
that $S^W_{A_i}\subset\{1,\ldots,K\}$ is an index set; if $S^W_{A_1} =
\{1,3,4\}$, then $W_{S^W_{A_i}} = (W_1, W_3, W_4).$ The same notion
applies to  $S^U_{A_i}\subset\{1,\ldots,J\}$.

The notation in this structural equation model is consistent with the
set up in \Cref{fig:general}. $W_k$'s are null confounders; $U_j$'s
are confounders; $V_l$'s are covariates. Moreover, $U_{S^U_{A_i}}$
indicates the set of confounders that have an arrow to both $A_i$ and
$Y$. $W_{S^W_{A_i}}$ indicates the set of null confounders that have
an arrow to $A_i$; they do not have arrows to $Y$.

Relating to the single-cause and multi-cause notion, we have
single-cause null confounders as
\begin{align}
\label{eq:singlecause}
W^{\mathrm{sng}}\stackrel{\Delta}{=}\{W_1,\ldots,
W_K\}/\bigcup_{i,j\in\{1,\ldots,m\}:i\ne j} (W_{S^W_{A_i}}\cap
W_{S^W_{A_j}}).
\end{align}
To parse the notation above, recall that $W_{S^W_{A_i}}$ is the set of
null confounders that affects $A_i$.
$\bigcup_{i,j\in\{1,\ldots,m\}:i\ne j} (W_{S^W_{A_i}}\cap
W_{S^W_{A_j}})$ describes the set of null confounders that affect at
least two of the $A_i$'s. Hence, $W^{\mathrm{sng}}$ denotes the set of
null confounders that affect only one of the $A_i$'s, a.k.a.
single-cause null confounders.

Before proving \Cref{lemma:reduce}, we first prove the following lemma
that quantifies the information in $Z$ (in \Cref{fig:generalredZ}).

\begin{lemma} 
\label{lemma:1}
The random variable $Z$ in \Cref{fig:generalredZ} ``captures'' all
multi-cause confounders, all multi-cause null confounders and some
independent error:
\begin{align}
\sigma(Z) &=\sigma\left(\{\epsilon_Z\} \bigcup (\cup_{i,j\in\{1,\ldots,m\}:i\ne
j} (W_{S^W_{A_i}}\cap W_{S^W_{A_j}})\cup(U_{S^U_{A_i}}\cap
U_{S^U_{A_j}}))\right),\\
& = \sigma\left(\{\epsilon_Z\} \bigcup
W^{\mathrm{mlt}}\bigcup U^{\mathrm{mlt}}\right).
\end{align}
where $\epsilon_Z\perp(\epsilon_Y, V_1, \ldots, V_L,
\cup_{i,j\in\{1,\ldots,m\}:i\ne j} (W_{S^W_{A_i}}\cap
W_{S^W_{A_j}})\cup(U_{S^U_{A_i}}\cap U_{S^U_{A_j}}), S)$.
\end{lemma}

We can parse the notation in \Cref{lemma:1} in the same way as in
\Cref{eq:singlecause}: $\cup_{i,j\in\{1,\ldots,m\}:i\ne
j}(W_{S^W_{A_i}}\cap W_{S^W_{A_j}})$ denotes the set of all
multi-cause confounders; $\cup_{i,j\in\{1,\ldots,m\}:i\ne j}
(U_{S^U_{A_i}}\cap U_{S^U_{A_j}})$ denotes the set of all multi-cause
null confounders.

\begin{proof}

Without the loss of generality, we assume the compactness of
representation in \Cref{eq:cause,eq:outcome}. For any subset
$\mathcal{S}$ of the random variables $\mathcal{S}
\subset\{A_1,\ldots,A_m, Y\},$ we assume the $\sigma$-algebra
$\sigma(\bigcap_{\tau}(S^W_{S_\tau}, S^U_{S_\tau}, S^V_{S_\tau}))$ is
the \emph{smallest} $\sigma$-algebra that makes all the random
variables in $\mathcal{S}$ jointly independent. The assumption is made
for technical convenience. We simply ensure the arrows from the $W, U,
V$'s to the $A_i$'s do exist. In other words, all the $W, U, V$'s
``whole-heartedly'' contribute to the $A_i$'s when they appear in
\Cref{eq:cause}. This assumption does not limit the class of causal
graphs we study.

First we show that all multi-cause confounders and all multi-cause
null confounders are measurable with respect to the substitute
confounder $Z$:
\begin{align}
\sigma\left(\bigcup_{i,j\in\{1,\ldots,m\}:i\ne
j} (W_{S^W_{A_i}}\cap W_{S^W_{A_j}})\cup(U_{S^U_{A_i}}\cap
U_{S^U_{A_j}})\right)\subset\sigma(Z).
\end{align}

Consider any pair of $A_i$ and $A_j$. \Cref{fig:generalredZ} implies
that
\begin{align}
\label{eq:paircondindep} A_i\perp A_j \g Z,
\end{align} for $i\ne j$ and $i,j\in\{1, \ldots, M\}.$ On the other
hand, we have
\begin{align} A_i\perp A_j \g \sigma\left((W_{S^W_{A_i}}\cap
W_{S^W_{A_j}}), (U_{S^U_{A_i}}\cap U_{S^U_{A_j}})\right),
\end{align} by the independence of errors assumption. Therefore, by
the compactness of representation assumption,
$\sigma((W_{S^W_{A_i}}\cap W_{S^W_{A_j}}), (U_{S^U_{A_i}}\cap
U_{S^U_{A_j}}))$ is the smallest $\sigma$-algebra that renders $A_i$
independent of $A_j$. This implies
\begin{align}
\sigma\left((W_{S^W_{A_i}}\cap
W_{S^W_{A_j}}), (U_{S^U_{A_i}}\cap
U_{S^U_{A_j}})\right)\subset\sigma(Z).
\end{align} The argument can be applied to any pair of $i\ne j,
i,j\in\{1, \ldots, M\}$, so we have
\begin{align}
\sigma\left(\bigcup_{i,j\in\{1,\ldots,m\}:i\ne
j} (W_{S^W_{A_i}}\cap W_{S^W_{A_j}})\cup(U_{S^U_{A_i}}\cap
U_{S^U_{A_j}})\right)\subset\sigma(Z)\label{eq:left}.
\end{align}
Next \Cref{fig:generalredZ} implies
\begin{align}
\sigma(A_1,\ldots, A_M)\not\subset\sigma(Z),
\end{align}
and
\begin{align}
\sigma(Y)\not\subset \sigma(Z).
\end{align}

Therefore, we have
\begin{align}
\label{eq:right}
\sigma(Z) \subset \sigma\left(\{\epsilon_Z\} \bigcup (\cup_{i,j\in\{1,\ldots,m\}:i\ne
j} (W_{S^W_{A_i}}\cap W_{S^W_{A_j}})\cup(U_{S^U_{A_i}}\cap
U_{S^U_{A_j}}))\right),
\end{align}
where $\epsilon_Z$ is independent of all the other errors in the
structural model, including those of $A$ and $Y$.

The error $\epsilon_Z$ can have an empty $\sigma$-algebra: for
example, $\epsilon_Z$ is a constant. Therefore, the left side of
\Cref{eq:left} can be made equal to the right side of \Cref{eq:right}.
We have
\begin{align}
\sigma(Z) &= \sigma\left(\{\epsilon_Z\} \bigcup
(\cup_{i,j\in\{1,\ldots,m\}:i\ne j} (W_{S^W_{A_i}}\cap
W_{S^W_{A_j}})\cup(U_{S^U_{A_i}}\cap U_{S^U_{A_j}}))\right)\\
& = \sigma\left(\{\epsilon_Z\} \bigcup
W^{\mathrm{mlt}}\bigcup U^{\mathrm{mlt}}\right).
\end{align}
for some random variable $\epsilon_Z$ that is independent of all other
random errors $\epsilon$'s.
\end{proof}

As a direct consequence of \Cref{lemma:1}, we have
\begin{align}
P(y\g u^{\mathrm{sng}}, u^{\mathrm{mlt}}, w^{\mathrm{mlt}}, a_1,
\ldots, a_m, s = 1)
=P(y\g u^{\mathrm{sng}}, z, a_1,
\ldots, a_m, s = 1),
\end{align}
due to the definition of conditional probabilities and
$\epsilon_Z\perp Y\g S, U^{\mathrm{sng}}, U^{\mathrm{mlt}},
W^{\mathrm{mlt}}, A_1, \ldots, A_m$. The latter is because
$\epsilon_Z$ is independent of all other errors.

\section{Proof of \Cref{lemma:idgeneralredZ}}
\label{sec:id-generalredZ-proof}

\begin{proof} Denote $U^{\mathrm{sng}}_{\mathcal{C}}$ as the set of
single-cause confounders that affects $A_\mathcal{C}.$

The proof of \Cref{lemma:idgeneralredZ} relies on two observations.

The first observation starts with the integral equation we solve:
\begin{align}
&P(y\g a_\mathcal{C}, f(a_\mathcal{N}), u^{\mathrm{sng}}_{\mathcal{C}}, s=1) \\
=&\int h(y, a_\mathcal{C},
a_\mathcal{X}, u^{\mathrm{sng}}_{\mathcal{C}}) P(a_\mathcal{X}\g a_\mathcal{C}, f(a_\mathcal{N}), u^{\mathrm{sng}}_{\mathcal{C}}, s=1)\dif a_\mathcal{X} \\
=&\int \int h(y, a_\mathcal{C},
a_\mathcal{X}, u^{\mathrm{sng}}_{\mathcal{C}}) P(a_\mathcal{X}\g z) P(z \g a_\mathcal{C}, f(a_\mathcal{N}), u^{\mathrm{sng}}_{\mathcal{C}}, s=1)\dif a_\mathcal{X}\dif z
\label{eq:obs1gen}
\end{align}

The first equality is due to \Cref{eq:integral-general}. The second equality
is due to \Cref{assumption:generalredZ}.2.

The second observation relies on the null proxy:
\begin{align}
&P(y\g a_\mathcal{C}, f(a_\mathcal{N}), u^{\mathrm{sng}}_{\mathcal{C}}, s=1) \\
=& \int P(y\g z, a_\mathcal{C}, f(a_\mathcal{N}), u^{\mathrm{sng}}_{\mathcal{C}}, s=1) P(z\g a_\mathcal{C}, f(a_\mathcal{N}), u^{\mathrm{sng}}_{\mathcal{C}}, s=1)\dif z\\
=& \int P(y\g z, a_\mathcal{C}, u^{\mathrm{sng}}_{\mathcal{C}}, s=1) P(z \g a_\mathcal{C}, f(a_\mathcal{N}), u^{\mathrm{sng}}_{\mathcal{C}}, s=1)\dif z\label{eq:obs2gen}
\end{align}
The first equality is due to the definition of conditional
probability. The second equality is due to the second part of
\Cref{assumption:null-proxy-generalredZ}; it implies $Y\perp
f(a_\mathcal{N})\g Z, U^{\mathrm{sng}}_{\mathcal{C}}, A_\mathcal{C},
S=1.$ The reason is that
\begin{align}
&P(y\g z, a_\mathcal{C}, f(a_\mathcal{N}), u^{\mathrm{sng}}_{\mathcal{C}}, s=1)\\
=&\int P(y\g z, a_\mathcal{C}, a_\mathcal{X}, f(a_\mathcal{N}), u^{\mathrm{sng}}_{\mathcal{C}}, s=1) P(a_\mathcal{X}\g z, a_\mathcal{C}, f(a_\mathcal{N}), u^{\mathrm{sng}}_{\mathcal{C}}, s=1)\dif a_\mathcal{X}\\
=&\int P(y\g z, a_\mathcal{C}, a_\mathcal{X}, u^{\mathrm{sng}}_{\mathcal{C}}, s=1) P(a_\mathcal{X}\g z, a_\mathcal{C},  u^{\mathrm{sng}}_{\mathcal{C}}, s=1)\dif a_\mathcal{X}\\
=& P(y\g z, a_\mathcal{C}, u^{\mathrm{sng}}_{\mathcal{C}}, s=1).
\end{align}
The second equality is again due to \Cref{assumption:generalredZ}.2.

Comparing \Cref{eq:obs1gen} and \Cref{eq:obs2gen} gives 
\begin{multline}
\int \left[P(y\g z, a_\mathcal{C}, u^{\mathrm{sng}}_{\mathcal{C}}, s=1)-\int h(y, a_\mathcal{C},
a_\mathcal{X}, u^{\mathrm{sng}}_{\mathcal{C}}) P(a_\mathcal{X}\g z)\dif a_\mathcal{X}\right]\\
 \times P(z\g a_\mathcal{C}, f(a_\mathcal{N}), u^{\mathrm{sng}}_{\mathcal{C}}, s=1)\dif z = 0,
\end{multline}
which implies
\begin{align}
\label{eq:fullcondgeneral}
P(y\g z, a_\mathcal{C}, u^{\mathrm{sng}}_{\mathcal{C}}, s=1) = \int h(y, a_\mathcal{C},
a_\mathcal{X}, u^{\mathrm{sng}}_{\mathcal{C}}) P(a_\mathcal{X}\g z)\dif a_\mathcal{X}.
\end{align}
This step is due to the completeness condition in
\Cref{assumption:null-proxy-generalredZ}.2.

\Cref{eq:fullcondgeneral} leads to identification:
\begin{align}
&P(y\g \mathrm{do}(a_\mathcal{C})) \\
=&P(y\g z, a_\mathcal{C}, u^{\mathrm{sng}}_{\mathcal{C}}) P(z) P(u^{\mathrm{sng}}_{\mathcal{C}})\dif z\dif u^{\mathrm{sng}}_{\mathcal{C}}\\
=&P(y\g z, a_\mathcal{C}, u^{\mathrm{sng}}_{\mathcal{C}}, s=1) P(z) P(u^{\mathrm{sng}}_{\mathcal{C}})\dif z\dif u^{\mathrm{sng}}_{\mathcal{C}}\\
=& \int \int \int h(y, a_\mathcal{C},
a_\mathcal{X}, u^{\mathrm{sng}}_{\mathcal{C}}) P(a_\mathcal{X}\g z)\dif a_\mathcal{X} P(z) P(u^{\mathrm{sng}}_{\mathcal{C}})\dif z \dif u^{\mathrm{sng}}_{\mathcal{C}}\\
=& \int \int h(y, a_\mathcal{C},
a_\mathcal{X}, u^{\mathrm{sng}}_{\mathcal{C}}) P(a_\mathcal{X}) P(u^{\mathrm{sng}}_{\mathcal{C}})\dif a_\mathcal{X} \dif u^{\mathrm{sng}}_{\mathcal{C}}.
\end{align}
In particular, the second equality is due to \Cref{assumption:generalredZ}.2.

\end{proof}

\section{Proof of \Cref{thm:idgeneral}}
\label{sec:id-general-proof}

We first state the variant of \Cref{assumption:generalredZ} and
\Cref{assumption:null-proxy-generalredZ}  required by
\Cref{thm:idgeneral}. We essentially replace $Z$ with
$(U^{\mathrm{mlt}}, W^{\mathrm{mlt}})$ in these assumptions.

\begin{assumption} (\Cref{assumption:generalredZ}')
\label{assumption:general}
The causal graph \Cref{fig:general} satisfies the following
conditions:
\begin{enumerate}
	\item All single-cause confounders $U^{\mathrm{sng}}_i$'s are observed.
	\item The selection operator $S$ satisfies
	\begin{align}
	S\perp (A, Y)\g  U^{\mathrm{mlt}}, W^{\mathrm{mlt}}, U^{\mathrm{sng}}.
	\end{align}
	\item We observe the non-selection-biased distribution \[P(a_1, \ldots, a_m,
u^{\mathrm{sng}})\] and the selection-biased distribution \[P(y,
u^{\mathrm{sng}}, a_1, \ldots, a_m \g s=1).\]
\end{enumerate}
\end{assumption}

\begin{assumption} (\Cref{assumption:null-proxy-generalredZ}')
\label{assumption:null-proxy-general}
There exists some function $f$ and a set
$\emptyset \ne \mathcal{N} \subset \{1, \ldots, m\}\backslash\mathcal{C}$
such that
% \vspace{-6pt}
\begin{enumerate}
\item The outcome $Y$ does not causally depend on $f(a_\mathcal{N})$:
\begin{align} 
\label{eq:nullproxyreqgeneral}
f(a_\mathcal{N})\perp Y \g A_\mathcal{C}, A_\mathcal{X}, U^{\mathrm{mlt}}, W^{\mathrm{mlt}}, U^{\mathrm{sng}}, S = 1
\end{align}
where $\mathcal{X}=  \{1, \ldots, m\}\backslash
(\mathcal{C}\cup \mathcal{N}) \ne \emptyset.$
\item The conditional $P(u^{\mathrm{mlt}}, w^{\mathrm{mlt}} \g a_\mathcal{C},
f(a_\mathcal{N}), u^{\mathrm{sng}}_\mathcal{C}, s=1)$ is complete in
$f(a_\mathcal{N})$  for almost all $a_\mathcal{C}$ and
$u^{\mathrm{sng}}_\mathcal{C}$, where $U^{\mathrm{sng}}_\mathcal{C}$
is the single-cause confounders affecting $A_\mathcal{C}$.
\item The conditional $P(f(a_\mathcal{N})\g
a_\mathcal{C}, a_\mathcal{X}, u^{\mathrm{sng}}_\mathcal{C}, s=1)$ is complete in $a_\mathcal{X}$ for
almost all $a_\mathcal{C}$ and $u^{\mathrm{sng}}_\mathcal{C}$.
\end{enumerate}
\end{assumption}

Under these assumptions, \Cref{thm:idgeneral} is a direct consequence
of \Cref{lemma:reduce} and \Cref{lemma:idgeneralredZ}. The reason is
that $U^{\mathrm{mlt}}, W^{\mathrm{mlt}}, U^{\mathrm{sng}}$
constitutes an admissible set to identify the intervention
distributions $P(y\g
\mathrm{do}(a_\mathcal{C}))$.

\section{Proof of \Cref{thm:dcfcorrect-general}}
\label{sec:dcfcorrect-general-proof}

We assume \Cref{assumption:general} and
\Cref{assumption:null-proxy-general} as described in
\Cref{sec:id-general-proof}.

\begin{proof} 

\Cref{assumption:dcf-generalZ}.2 guarantees the existence of some
function $\hat{h}$ such that
\begin{align}
\label{eq:dcfgeneral2supp}
\hat{P}(y\g a_\mathcal{C},
\hat{z}, u^{\mathrm{sng}}_\mathcal{C}, s = 1) =\int \hat{h}(y, a_\mathcal{C},
a_\mathcal{X}, u^{\mathrm{sng}}_\mathcal{C}) \hat{P}(a_\mathcal{X}\g \hat{z}, u^{\mathrm{sng}}_\mathcal{C}, s = 1)\dif a_\mathcal{X}
\end{align}
under weak regularity conditions. (We discuss the reason in
\Cref{sec:integral-sol-exist}.)

We first claim that $\hat{h}(y, a_\mathcal{C},
a_\mathcal{X}, u^{\mathrm{sng}}_\mathcal{C})$ solves
\begin{align}
\label{eq:keyeq}
P(y\g a_\mathcal{C}, f(a_\mathcal{N}), u^{\mathrm{sng}}_{\mathcal{C}}, s=1)
=& \int \hat{h}(y, a_\mathcal{C},
a_\mathcal{X}, u^{\mathrm{sng}}_{\mathcal{C}}) P(a_\mathcal{X}\g a_\mathcal{C}, f(a_\mathcal{N}), u^{\mathrm{sng}}_{\mathcal{C}}, s=1)\dif a_\mathcal{X}.
\end{align}

Given this claim (\Cref{eq:keyeq}), we have
\begin{align*}
&\hat{P}(y\g \mathrm{do}(a_\mathcal{C})) \\
=& \int \int \hat{P}(y\g \hat{z},u^{\mathrm{sng}}_{\mathcal{C}},
a_\mathcal{C}, s=1)\hat{P}(\hat{z})P(u^{\mathrm{sng}}_{\mathcal{C}})\dif \hat{z}\dif u^{\mathrm{sng}}_{\mathcal{C}}\\
=& \int \int \int \hat{h}(y, a_\mathcal{C},
a_\mathcal{X}, u^{\mathrm{sng}}_\mathcal{C}) \hat{P}(a_\mathcal{X}\g \hat{z}, u^{\mathrm{sng}}_\mathcal{C}, s = 1)\dif a_\mathcal{X}\hat{P}(\hat{z})P(u^{\mathrm{sng}}_{\mathcal{C}})\dif \hat{z}\dif u^{\mathrm{sng}}_{\mathcal{C}}\\
=& \int \int \int \hat{h}(y, a_\mathcal{C},
a_\mathcal{X}, u^{\mathrm{sng}}_\mathcal{C}) \hat{P}(a_\mathcal{X}\g \hat{z})\dif a_\mathcal{X}\hat{P}(\hat{z})P(u^{\mathrm{sng}}_{\mathcal{C}})\dif \hat{z}\dif u^{\mathrm{sng}}_{\mathcal{C}}\\
=& \int \int  \hat{h}(y, a_\mathcal{C},
a_\mathcal{X}, u^{\mathrm{sng}}_\mathcal{C}) P(a_\mathcal{X})\dif a_\mathcal{X}P(u^{\mathrm{sng}}_{\mathcal{C}})\dif u^{\mathrm{sng}}_{\mathcal{C}}\\
=& P(y\g \mathrm{do}(a_\mathcal{C})),
\end{align*}
which proves the theorem. The first equality is due to
\Cref{eq:dcfgeneral}; the second is due to \Cref{eq:dcfgeneral2supp};
the third is due to \Cref{assumption:dcf-generalZ} and
$U^{\mathrm{sng}}_{\mathcal{C}}$ being the single-cause confounders
for $A_\mathcal{C}$; the fourth is due to marginalizing out $\hat{Z}$;
the fifth is due to the above claim (\Cref{eq:keyeq}) and
\Cref{thm:idgeneral}.

We next prove the claim (\Cref{eq:keyeq}). Start with the right side of
the equality.
\begin{align*}
&\int \hat{h}(y, a_\mathcal{C},
a_\mathcal{X}, u^{\mathrm{sng}}_{\mathcal{C}}) P(a_\mathcal{X}\g a_\mathcal{C}, f(a_\mathcal{N}), u^{\mathrm{sng}}_{\mathcal{C}}, s=1)\dif a_\mathcal{X}\\
=& \int \int \hat{h}(y, a_\mathcal{C},
a_\mathcal{X}, u^{\mathrm{sng}}_{\mathcal{C}})  \hat{P}(a_\mathcal{X}\g \hat{z},u^{\mathrm{sng}}_{\mathcal{C}},
a_\mathcal{C}, s=1 ) \hat{P}(\hat{z}\g a_\mathcal{C}, f(a_\mathcal{N}), u^{\mathrm{sng}}_{\mathcal{C}}, s=1)\dif a_\mathcal{X} \dif \hat{z}\\
=& \int \hat{P}(y\g a_\mathcal{C},
\hat{z}, u^{\mathrm{sng}}_\mathcal{C}, s = 1) \hat{P}(\hat{z}\g a_\mathcal{C}, f(a_\mathcal{N}), u^{\mathrm{sng}}_{\mathcal{C}}, s=1)\dif \hat{z}\\
=& \int \hat{P}(y\g a_\mathcal{C}, f(a_\mathcal{N}),
\hat{z}, u^{\mathrm{sng}}_\mathcal{C}, s = 1) \hat{P}(\hat{z}\g a_\mathcal{C}, f(a_\mathcal{N}), u^{\mathrm{sng}}_{\mathcal{C}}, s=1)\dif \hat{z}\\
=& P(y\g a_\mathcal{C}, f(a_\mathcal{N}), u^{\mathrm{sng}}_{\mathcal{C}}, s=1),
\end{align*}

which establishes the claim. The first equality is due to
\Cref{eq:dcfgeneralreq}; the second is due to
\Cref{eq:dcfgeneral2supp}; the third equality is due to
\Cref{assumption:dcf-generalZ}.2, which implies
\begin{align} 
\hat{P}(y\g a_\mathcal{C}, f(a_\mathcal{N}), 
\hat{z}, u^{\mathrm{sng}}_\mathcal{C}, s = 1) = \hat{P}(y\g a_\mathcal{C},
\hat{z}, u^{\mathrm{sng}}_\mathcal{C}, s = 1).
\end{align}
The fourth equality is due to marginalizing out $\hat{z}$.
\end{proof}

\section{Constructing candidate $f(a_\mathcal{N})$'s from the
deconfounder outcome model}

\label{sec:construct-fAN}

We illustrate how to construct candidate $f(a_\mathcal{N})$'s in the
deconfounder outcome model.

Consider a fitted linear outcome model
\begin{align}
Y = \sum_{i=1}^{10}\alpha_{YA_i} A_i + \alpha_{YZ}\hat{Z} + \alpha_{YU'}U^\mathrm{sng}+\epsilon_{Y}.
\end{align}
where all the random variables are Gaussian.

It implies that there exists $f_1(A_9, A_{10}) = A_9 + \alpha_{9,
10}A_{10}$ that satisfies
\[f_1(A_9, A_{10})\perp Y\g \hat{Z}, U^\mathrm{sng}, A_1, \ldots, A_8,\] where
\[\alpha_{9,
10}=-\frac{\alpha_9\mathrm{Var}(A_9)+\alpha_{10}\mathrm{Cov}(A_9,
A_{10})}{\alpha_9\mathrm{Cov}(A_9,
A_{10})+\alpha_{10}\mathrm{Var}(A_{10})}.\] The reason is that $f(A_9,
A_{10})\perp (\alpha_9A_9+\alpha_{10}A_{10})$. Hence $f(a_\mathcal{N})
= A_9 + \alpha_{9, 10}A_{10}$ satisfies
\Cref{assumption:dcf-generalZ}.2.

%%% Local Variables:
%%% mode: latex
%%% TeX-master: "dcf_SEM"
%%% End:
}
\clearpage
\putbib[BIB1]
\end{bibunit}

\end{document}